%% file: main.tex
\documentclass[10pt,twocolumn,letterpaper]{article}

\usepackage{wacv}
\usepackage{times}
\usepackage{epsfig}
\usepackage{graphicx}
\usepackage{amsmath}
\usepackage{amssymb}
\usepackage{subfigure}
\usepackage{xspace}

\usepackage{microtype}

\usepackage{xcolor}
\usepackage{textcomp}
\usepackage{color}
\usepackage{colortbl}
\usepackage{url}
\usepackage{gensymb}
\usepackage{multirow}

\graphicspath{{./graphics/}}
\usepackage{animate}
\usepackage{tabularx}
\usepackage{booktabs}
\usepackage{caption}
\usepackage{tikz}
\usepackage{pgfplots}
\usepackage{pgfplotstable}
\usepackage[skins]{tcolorbox}
\usepackage{standalone}
\usepackage{bm}
\usepackage{mleftright}
\usepackage{nicefrac}
\usepackage{makecell}
\usepackage{textcomp}
\usepackage[normalem]{ulem}
\usepackage[shortcuts]{extdash}
\usepackage[outline]{contour}
\newlength{\itemwidth}

\newcolumntype{Y}{>{\centering\arraybackslash}X}
\newcolumntype{P}[1]{>{\centering\arraybackslash}p{#1}}

\usetikzlibrary{calc}
\usetikzlibrary{tikzmark}
\usetikzlibrary{spy}
\pgfplotsset{compat=newest}

\definecolor{619D47}{RGB}{97,157,71}
\definecolor{EE7F0E}{RGB}{238,127,14}
\definecolor{3787CF}{RGB}{55,135,207}
\definecolor{DBDC4A}{RGB}{219,220,74}
\definecolor{3787CF}{RGB}{55,135,207}

\pgfplotscreateplotcyclelist{custompalette}{
    {619D47!90!black,fill=619D47},
    {EE7F0E!90!black,fill=EE7F0E},
    {3787CF!90!black,fill=3787CF},
    {DBDC4A!90!black,fill=DBDC4A}
}
\pgfplotscreateplotcyclelist{anothercustompalette}{
    {619D47!90!black,line width=1.5pt},
    {EE7F0E!90!black,line width=1.5pt},
    {3787CF!90!black,line width=1.5pt},
    {DBDC4A!90!black,line width=1.5pt}
}

\usepackage{stmaryrd}
\usepackage{trimclip}

\makeatletter
\DeclareRobustCommand{\shortto}{%
  \mathrel{\mathpalette\short@to\relax}%
}

\DeclareRobustCommand{\veryshortto}{%
  \mathrel{\mathpalette\veryshort@to\relax}%
}

\newcommand{\short@to}[2]{%
  \mkern2mu
  \clipbox{{.3\width} 0 0 0}{$\m@th#1\vphantom{+}{\shortrightarrow}$}%
  }

\newcommand{\veryshort@to}[2]{%
  \mkern2mu
  \clipbox{{.2\width} 0 0 0}{$\m@th#1\vphantom{+}{\shortrightarrow}$}%
  }
\makeatother

\tikzset{
  double arrow/.style args={#1 with #2 and #3}{
    -stealth, line width=#1, #2, postaction={
        draw, -stealth, line width=(#1)/2, shorten <= (#1)/4, shorten >= 2*(#1)/4, #3
    }
  }
}

\tikzset{
  double arrow bothdir/.style args={#1 with #2 and #3}{
    stealth-stealth, line width=#1, #2, postaction={
        draw, stealth-stealth, line width=(#1)/2, shorten <= 2*(#1)/4, shorten >= 2*(#1)/4, #3
    }
  }
}

\wacvfinalcopy

\usepackage[pagebackref=true,breaklinks=true,colorlinks,bookmarks=false]{hyperref}

\begin{document}

\title{\vspace{-0.6cm}Learned Dual-View Reflection Removal\vspace{-0.1cm}}

\makeatletter
\g@addto@macro\@maketitle{
    \vspace*{-15pt}
    \begin{center}\centering
        \setlength{\tabcolsep}{0.05cm}
        \setlength{\itemwidth}{3.41cm}
        \hspace*{-\tabcolsep}\begin{tabular}{ccccc}
               \includegraphics[width=\itemwidth, trim={0.0cm 0.0cm 0.0cm 1.0cm}, clip]{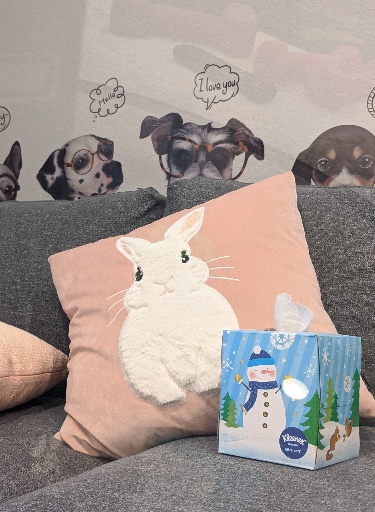}
            &
               \includegraphics[width=\itemwidth, trim={0.0cm 0.0cm 0.0cm 1.0cm}, clip]{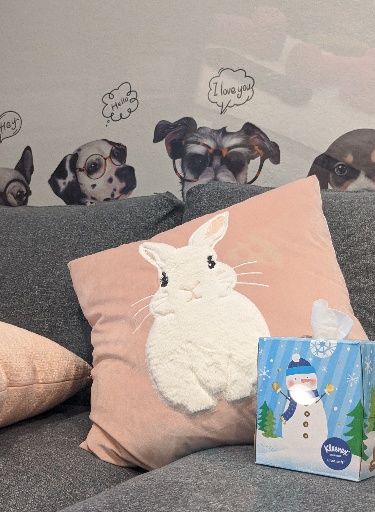}
            &
               \includegraphics[width=\itemwidth, trim={0.0cm 0.0cm 0.0cm 1.0cm}, clip]{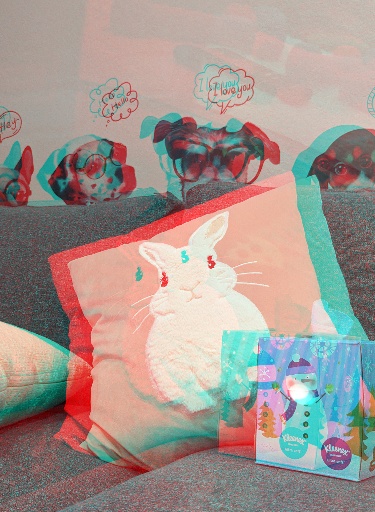}
            &
                \includegraphics[width=\itemwidth, trim={0.0cm 0.0cm 0.0cm 1.0cm}, clip]{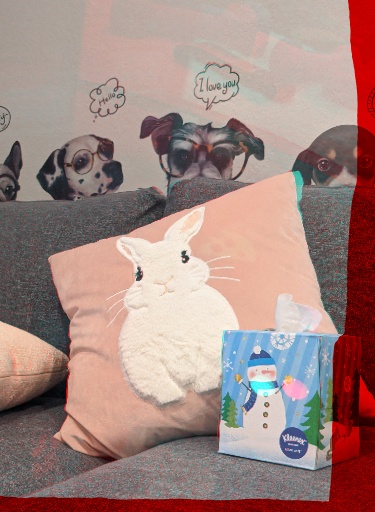}
            &
                \includegraphics[width=\itemwidth, trim={0.0cm 0.0cm 0.0cm 1.0cm}, clip]{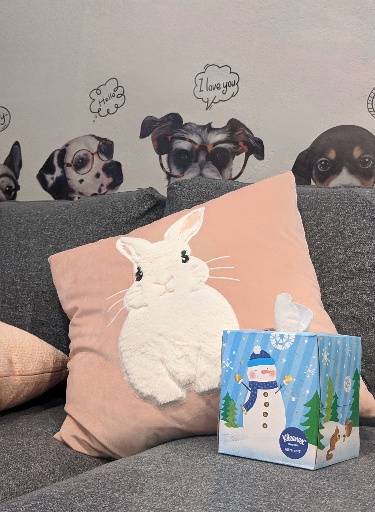}
            \\
                \footnotesize (a) View 1
            &
                \footnotesize (b) View 2
            &
                \footnotesize (c) Anaglyph
            &
                \footnotesize (d) Aligned
            &
                \footnotesize (e) Ours
            \\
        \end{tabular}\vspace{-0.2cm}
    	\captionof{figure}{Stereo pairs (a, b) were imaged through glass and exhibit undesired reflections. The transmitted and reflective images are subject to parallax that is difficult to separate as shown in the anaglyph (c). Our reflection-invariant flow aligns the two views with respect to the transmitted image, causing all remaining parallax (in the reflection on the tissue box, for example) to be due to reflections as shown in anaglyph (d). Our synthesis network exploits this parallax to remove reflections (e).}\vspace{-0.0cm}
    	\label{fig:teaser}
    \end{center}
}
\makeatother

\author{
\and
\and
Simon Niklaus\\
{\small Adobe Research}
\and
Xuaner (Cecilia) Zhang\\
{\small UC Berkeley}
\and
Jonathan T. Barron\\
{\small Google Research}
\and
\and
\and
Neal Wadhwa\\
{\small Google Research}
\and
Rahul Garg\\
{\small Google Research}
\and
Feng Liu\\
{\small Portland State University}
\and
Tianfan Xue\\
{\small Google Research}
}

\maketitle
\thispagestyle{empty}

\begin{abstract}

    Traditional reflection removal algorithms either use a single image as input, which suffers from intrinsic ambiguities, or use multiple images from a moving camera, which is inconvenient for users. We instead propose a learning-based dereflection algorithm that uses stereo images as input. This is an effective trade-off between the two extremes: the parallax between two views provides cues to remove reflections, and two views are easy to capture due to the adoption of stereo cameras in smartphones. Our model consists of a learning-based reflection-invariant flow model for dual-view registration, and a learned synthesis model for combining aligned image pairs. Because no dataset for dual-view reflection removal exists, we render a synthetic dataset of dual-views with and without reflections for use in training. Our evaluation on an additional real-world dataset of stereo pairs shows that our algorithm outperforms existing single-image and multi-image dereflection approaches. {\let\thefootnote\relax\footnote{\hspace{-0.5cm} Work primarily done while Simon and Xuaner were interns at Google.}}

\end{abstract}

\vspace*{-0.3cm}
\section{Introduction}
\label{sec:intro}
\input{intro}

\section{Related Work}
\label{sec:related}
\input{related}

\section{Method}
\label{sec:method}
\input{method}

\section{Experiments}
\label{sec:experiments}
\input{experiments}

\section{Conclusion}
\label{sec:conclusion}

In this paper, we presented a new learning-based dual-view reflection removal approach. Unlike the traditional reflection removal techniques, which either take a single frame or multiple frames as input, we proposed to use dual-view inputs, which yields a nice trade-off between the convenience of capturing and the resulting quality. To train this learned dual-view dereflection approach, we created a new dual-view  dataset by rendering realistic virtual environments. We also designed a new composite network consisting of a reflection-invariant optical flow estimation network and a dual-view transmission synthesis network. We have shown promising experimental results on both synthetic and real images with challenging reflections, outperforming previous work.

{\small
\bibliographystyle{ieee_fullname}
\bibliography{main}
}

\end{document}

%% file: intro.tex
Of the billions of pictures taken every year, a significant portion are taken through a reflective surface such as a glass window of a car or a glass case in a museum. This presents a problem for the photographer, as glass reflects some of the incident light from the same side as the photographer back towards the camera, corrupting the captured images with reflected image content. Formally, the captured image $I$ is the sum of the image being transmitted through the glass $T$ and the image of the light being reflected by the glass $R$:
\begin{equation}
    I[x, y, c] = T[x,y,c] + R[x,y,c]. \label{eq:formation}
\end{equation}
The task of reflection removal is estimating the image $T$ from an input image $I$. A solution to this problem has significant value, as it would greatly broaden the variety of circumstances in which photography can occur.

Equation~\ref{eq:formation} shows the core difficulty of single-image reflection removal: the problem is inherently underconstrained, as we have six unknowns at each pixel but only three observations. Most single-image techniques for reflection removal try to mitigate this problem by using image priors to disambiguate between reflection and transmission. Despite significant progress, most algorithms still cannot cleanly separate them. In fact, even humans may have difficulty when just given a single image. For example, it is difficult to tell whether the white spot next to the snowman in Figure~\ref{fig:teaser}(a) is a reflection or not without having a second perspective.

\begin{figure}\centering
    \setlength{\tabcolsep}{0.05cm}
    \setlength{\itemwidth}{4.11cm}
    \hspace*{-\tabcolsep}\begin{tabular}{cc}
            \begin{tikzpicture}
                \definecolor{arrowcolor}{RGB}{238,127,14}
                \node [anchor=south west, inner sep=0.0cm] (image) at (0,0) {
                    \includegraphics[width=\itemwidth, trim={0.0cm 1.0cm 0.0cm 0.0cm}, clip]{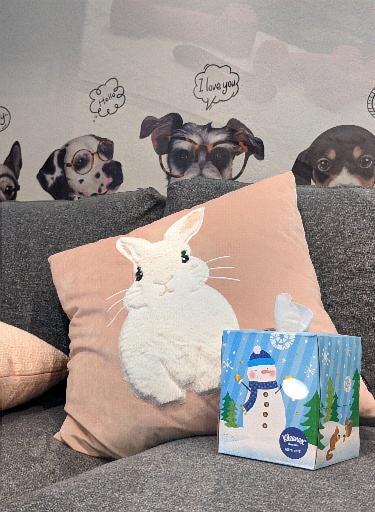}
                };
                \begin{scope}[x={(image.south east)},y={(image.north west)}]
                    \draw [double arrow=0.2cm with white and arrowcolor] (0.27,0.67) -- (0.37,0.87);
                    \draw [double arrow=0.2cm with white and arrowcolor] (0.44,0.07) -- (0.7,0.17);
                \end{scope}
            \end{tikzpicture}
            &
            \begin{tikzpicture}
                \definecolor{arrowcolor}{RGB}{97,157,71}
                \node [anchor=south west, inner sep=0.0cm] (image) at (0,0) {
                    \includegraphics[width=\itemwidth, trim={0.0cm 1.0cm 0.0cm 0.0cm}, clip]{graphics/motivation/ours-dual-lossy}
                };
                \begin{scope}[x={(image.south east)},y={(image.north west)}]
                    \draw [double arrow=0.2cm with white and arrowcolor] (0.27,0.67) -- (0.37,0.87);
                    \draw [double arrow=0.2cm with white and arrowcolor] (0.44,0.07) -- (0.7,0.17);
                \end{scope}
            \end{tikzpicture}
        \\
            \footnotesize (a) Ablation - Using One View
        &
            \footnotesize (b) Ours - Using Two Views
        \\
    \end{tabular}\vspace{-0.2cm}
	\caption{Comparison of a single-view ablation (a) to our proposed dual-view reflection removal (b). Reasoning jointly about both views allows our proposed approach to handle challenging scenes like this one. In comparison, the single-view ablation fails to remove all present reflections due to the underconstrained nature of the single-image setting.}\vspace{-0.5cm}
	\label{fig:motodual}
\end{figure}

The ambiguity of the single-image case led to the development of multi-image techniques. Figure~\ref{fig:teaser}(a) and \ref{fig:teaser}(b) show two views of a scene in which the camera translates slightly. Because the reflective and transmissive layers do not have the same distance from the camera, the scene content of the reflective layer moves differently from the transmissive layer when switching between the two views as shown in Figure~\ref{fig:teaser}(c). This parallax can help to disambiguate between reflection and transmission, thereby simplifying the task of recovering the constituent images. For this reason, practical systems for reflection removal rely on acquiring many images or entire videos of the same subject under different viewpoints~\cite{Liu_OTHER_2017, Xue_TOG_2015}. However, this setup is burdensome as it requires users to manually move their camera while capturing many images, and it assumes a static scene.

\begin{figure}\centering
    \setlength{\tabcolsep}{0.05cm}
    \setlength{\itemwidth}{4.11cm}
    \hspace*{-\tabcolsep}\begin{tabular}{cc}
            \begin{tikzpicture}[spy using outlines={circle, 3787CF, magnification=3, size=2.5cm, connect spies,
    every spy in node/.append style={line width=0.06cm}}]
                \node [inner sep=0.0cm] {\includegraphics[width=\itemwidth, interpolate=true, trim={0.0cm 1.0cm 0.0cm 0.0cm}, clip]{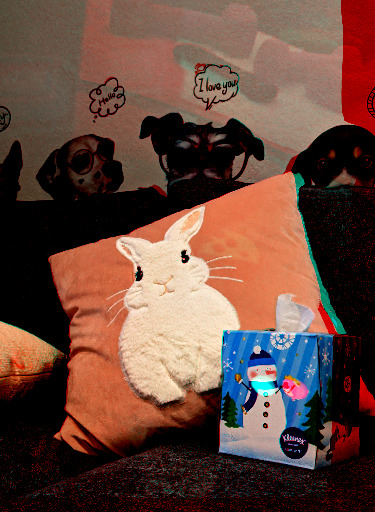}};
                \spy [every spy on node/.append style={line width=0.06cm}, spy connection path={\draw[line width=0.06cm] (tikzspyonnode) -- (tikzspyinnode);}] on (0.7,2.05) in node at (-0.6,-1.18);
            \end{tikzpicture}
        &
            \begin{tikzpicture}[spy using outlines={circle, 3787CF, magnification=3, size=2.5cm, connect spies,
    every spy in node/.append style={line width=0.06cm}}]
                \node [inner sep=0.0cm] {\includegraphics[width=\itemwidth, interpolate=true, trim={0.0cm 1.0cm 0.0cm 0.0cm}, clip]{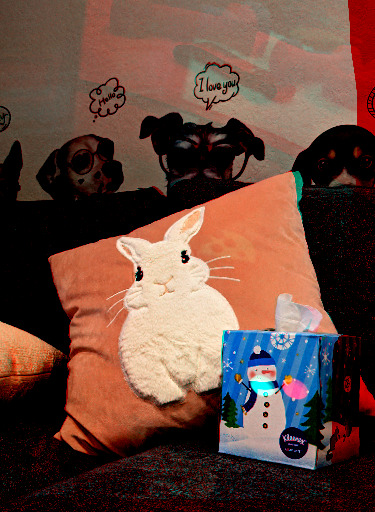}};
                \spy [every spy on node/.append style={line width=0.06cm}, spy connection path={\draw[line width=0.06cm] (tikzspyonnode) -- (tikzspyinnode);}] on (0.7,2.05) in node at (-0.6,-1.18);
            \end{tikzpicture}
        \\
            \footnotesize (a) Traditional Optical Flow
        &
            \footnotesize (b) Reflection-Invariant Flow
        \\
    \end{tabular}\vspace{-0.2cm}
	\caption{Aligned stereo anaglyphs by warping $I_2$ to $I_1$ with traditional optical flow (a), and our reflection-invariant optical flow (b). Contrast adjusted for visualization. Traditional flow aligns all image content, minimizing the parallax in both transmission and reflection. With our reflection-invariant optical flow, all remaining parallax is in the reflection.}\vspace{-0.3cm}
	\label{fig:motoflow}
\end{figure}

This points to a fundamental tension between single-image and multi-image techniques. We explore a compromising solution in which we take as input two views of the same scene produced by a stereo camera (Figure~\ref{fig:motodual}). 
Though binocular stereo is not new, smartphones are adopting camera arrays, thereby increasing the practicality of algorithms designed for stereo images. This presents an opportunity for high-quality dual-view dereflection that is as convenient as any single-image technique, requiring just a single button press and being capable of capturing non-static scenes.

Still, it is not trivial to extend existing single- or multi-image dereflection algorithms to dual-view input. Most multi-image algorithms~\cite{Xue_TOG_2015, Yu_OTHER_2013} use hand-tuned heuristics based on motion parallax and require at least 3 to 5 frames as input, as two views are often not enough to make this problem well-posed. And most single-image dereflection algorithms~\cite{Fan_ICCV_2017, Kim_CVPR_2020, Wen_CVPR_2019, Xuzhang_CVPR_2018} are trained on images with synthetic reflections, a strategy which does not generalize to dual-view input due to the need for realistic motion parallax.

To address these issues, we combine merits of both approaches and propose a learned approach that utilizes motion parallax. We first align the two input images using the motion of only the transmissive layer. Ignoring reflective content during registration produces aligned images where the transmissive layer is static while the reflection ``moves'' across aligned views, reducing the transmission-reflection separation problem to one of simply distinguishing between static and moving edges, as shown in Figure~\ref{fig:motoflow}(b). Unlike traditional flow approaches, which align both transmissive and reflective image content as shown in Figure~\ref{fig:motoflow}(a), we explicitly train an optical flow network to be invariant to reflections. After performing this reflection-invariant alignment, we supervise a image synthesis network to recover the transmission from the transmission-aligned views.

While this framework is conceptually simple, training such a model requires difficult-to-acquire dual-view imagery that is subject to reflections. It is even more difficult to obtain such data with accurate ground truth optical flow of the transmissive layer. As such, we resort to employing computer graphics and render virtual environments to create such a dataset. We also collect a real-world dual-view dataset with ground truth transmission for evaluation purposes, and show that our approach generalizes well to this data.

%% file: related.tex
\begin{figure*}\centering
    \includegraphics[]{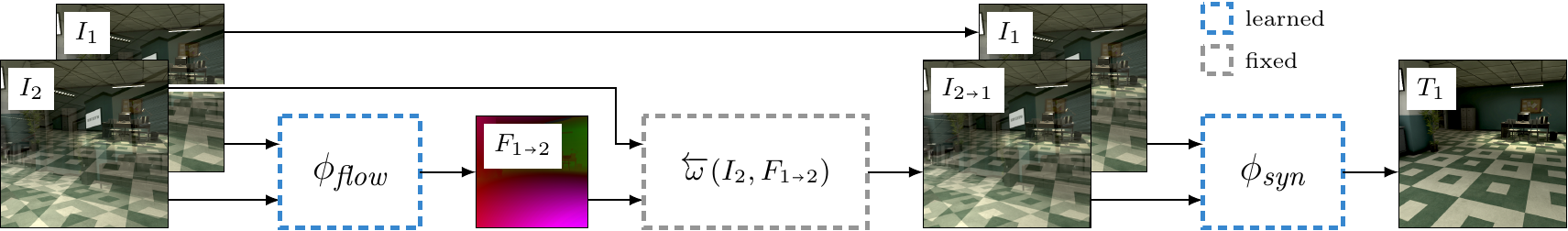}\vspace{-0.1cm}
	\caption{Our dual-view reflection removal. Given images $I_1$ and $I_2$, our reflection-invariant optical flow network $\phi_{flow}$ estimates the motion $F_{1 \shortto 2}$ between the unknown transmissive layers of the inputs, which is then used to warp $I_2$ towards $I_1$ to ``undo'' that motion. Our synthesis network $\phi_{syn}$ can then use these aligned images to leverage the parallax between the reflective layers (and the lack thereof between the transmissive layers) to synthesize $I_1$'s transmissive layer $T_1$.}\vspace{-0.4cm}
	\label{fig:architecture}
\end{figure*}

The task of reflection removal is a narrow sub-problem of the classical problem of inferring a complete model of the physical world that generated an observed image~\cite{Barrow_OTHER_1978}, which has been extensively studied throughout the history of computer vision. Reflection removal is similar in nature to other blind signal separation problems in computer vision, such as disentangling reflectance and shading~\cite{Barron_PAMI_2015} or separating haze from transmitted scene content~\cite{He_PAMI_2011}. Due to the ill-posed nature of reflection removal, many past works used additional information to constrain the problem. A common strategy is to use multiple images captured from different viewpoints as input, taking advantage of how transmitted content is constant across images while the reflective content changes~\cite{Han_CVPR_2017, Li_ICCV_2013, Liu_OTHER_2017, Xue_TOG_2015}. These approaches require significant labor from the photographer, and also assume a static scene. Another approach is to use multiple images from the same view but with different polarization~\cite{Kong_PAMI_2014, Schechner_ICCV_1999}, which leverages the relationship between the angle of incidence of light on the reflecting surface and its polarization. Though effective, these techniques require a static scene and the rather exotic ability to modify a camera's polarization.

Automatic single-image reflection removal techniques are an attractive alternative to multi-image solutions~\cite{Wan_ICCV_2017}. Prior to the rise of deep learning, single-image reflection techniques would usually impose beliefs about the natural world or the appearance of reflected images, and then recover the transmittance and reflectance that best satisfy those priors. These approaches require the manual construction of regularizers on edges or relative smoothness~\cite{Levin_CVPR_2004, Li_CVPR_2014, Shih_CVPR_2015, Yang_CVPR_2019}, then solving an expensive and/or non-convex optimization problem. With deep learning, the focus shifted towards training a network to map from the input image to the transmission~\cite{Fan_ICCV_2017, Li_CVPR_2020, Wan_CVPR_2018, Yang_ECCV_2018, Xuzhang_CVPR_2018}. Though effective, these techniques depend critically on the quality of training data.

Our work addresses an unexplored approach that lies between single-image and multi-image cases. By combining the information present in stereo imagery with the effectiveness of a neural network trained on vast amounts of synthetic data, our approach produces higher-quality output than single-image approaches while requiring none of the labor or difficulty of multi-image approaches.

Stereo cameras are closely related to dual-pixel sensors, wherein a single camera has a sensor with ``split'' pixels, thereby allowing it to produce limited light fields~\cite{Garg_ICCV_2019, Wadhwa_TOG_2018}.
Dual-pixel reflection removal has been explored with promising results~\cite{Punnappurath_CVPR_2019}, but it is unclear how such a technique might generalize to stereo. First, the dual-pixel disparity is only significant in cameras with large apertures, like DSLRs but not smartphones. When using a DSLR though, reflections are out of focus and are heavily blurred which in itself already provides important cues. Second, due to the interplay between focus distance and dual-pixel images, one can simply threshold the dual-pixel disparity to separate reflection edges from transmitted content as done in~\cite{Punnappurath_CVPR_2019}. Such a universal threshold does unfortunately not exist for stereo images.

%% file: method.tex
Given images $I_1$ and $I_2$ captured from two different viewpoints, our goal is to estimate $T_1$, an image that contains only the transmissive content of $I_1$. We have found that a single network is unable to synthesize $T_1$ from $I_1$ and $I_2$ directly, presumably due to the difficulty of simultaneously aligning and combining these images. We hence decompose this task into: reflection-invariant motion estimation, warping to account for transmission parallax, and transmission synthesis. We recover the optical flow $F_{1 \shortto 2}$ between the transmissive layers of $I_1$ and $I_2$ using a network $\phi_{flow}$ as
\begin{equation}
    F_{1 \shortto 2} = \phi_{flow} \left( I_1, I_2 \right)
\label{eqn:flow}
\end{equation}
This step depends critically on $\phi_{flow}$ being trained to be invariant to reflection, as we describe in Section~\ref{sec:invariant_flow}. We then use this optical flow to account for the inter-frame transmission motion via differentiable sampling~\cite{Jaderberg_NIPS_2015}. Specifically, we use backward warping $\overleftarrow{\omega}$ and warp $I_2$ to $I_1$ according to the estimated optical flow $F_{1 \shortto 2}$ to generate $I_{2 \shortto 1}$ as
\begin{equation}
    I_{2 \shortto 1} = \overleftarrow{\omega} \left( I_2, F_{1 \shortto 2} \right),
\label{eqn:warp}
\end{equation}
Because our optical flow is reflection-invariant, $I_2$ is warped such that only its transmissive content matches that of $I_1$. This allows us to apply a synthesis model that takes as input the image of interest $I_1$ and its warped counterpart $I_{2 \shortto 1}$, and estimates the first image's transmissive layer $T_1$ as
\begin{equation}
    T_1 = \phi_{syn} \left( I_1, I_{2 \shortto 1}\right).
\label{eqn:sync}
\end{equation}
Combining these Equations~\ref{eqn:flow}--\ref{eqn:sync} gives our complete reflection removal pipeline, which we also visually summarize in  Figure~\ref{fig:architecture}, where $\phi_{flow}$ and $\phi_{syn}$ are neural networks.

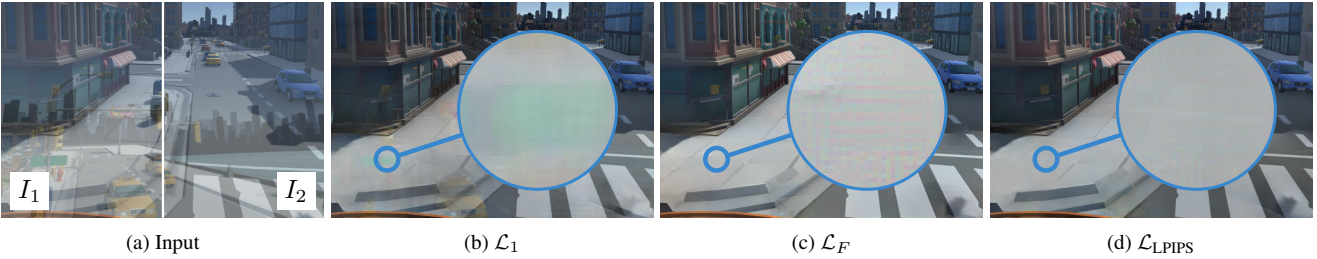
\begin{figure*}\centering
    \setlength{\tabcolsep}{0.05cm}
    \setlength{\itemwidth}{4.28cm}
    \hspace*{-\tabcolsep}\begin{tabular}{cccc}
            \begin{tikzpicture}[scale=4.28]\begin{scope}\clip (0,0) rectangle (1,0.668);
                \begin{scope}
                    \clip (0,1) -- (0.5,1) -- (0.5,0) -- (0,0) -- cycle;
                    \fill [fill overzoom image=graphics/loss/input-1] (0,0) rectangle (1,1);
                    \node [anchor=south west, fill=white, inner sep=0.1cm] at (0.025,0.025) {$I_1$};
                \end{scope}
                \begin{scope}
                    \clip (1,1) -- (0.5,1) -- (0.5,0) -- (1,0) -- cycle;
                    \fill [fill overzoom image=graphics/loss/input-2] (0,0) rectangle (1,1);
                    \node [anchor=south east, fill=white, inner sep=0.1cm] at (0.975,0.025) {$I_2$};
                \end{scope}
                \draw [white, line width=0.03cm] (0.5,0) -- (0.5,1.0);
                \node [inner sep=0.0cm] at (0.5,0.5*0.668) {
                    \begin{animateinline}[autoplay, palindrome, final, nomouse, method=widget, poster=none]{0.5}
                        \begin{tikzpicture}
                            \node [anchor=south west, inner sep=0.0cm] (image) at (0,0) {
                                \includegraphics[width=\itemwidth, trim={0.0cm 0.0cm 0.0cm 6.0cm}, clip]{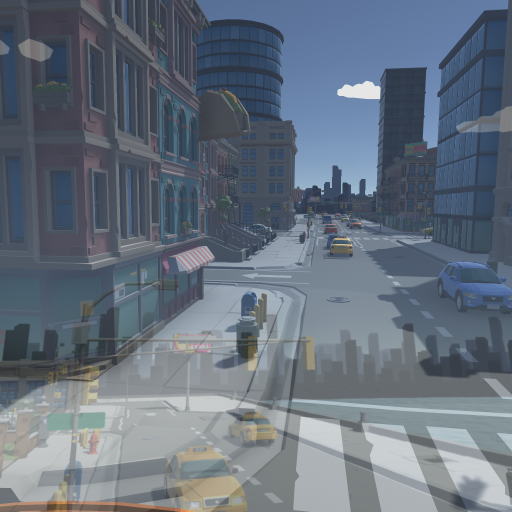}
                            };
                            \begin{scope}[x={(image.south east)},y={(image.north west)}]
                                \node [anchor=south west, fill=white, inner sep=0.1cm] at (0.025,0.035) {$I_1$};
                            \end{scope}
                        \end{tikzpicture}
                        \newframe
                        \begin{tikzpicture}
                            \node [anchor=south west, inner sep=0.0cm] (image) at (0,0) {
                                \includegraphics[width=\itemwidth, trim={0.0cm 0.0cm 0.0cm 6.0cm}, clip]{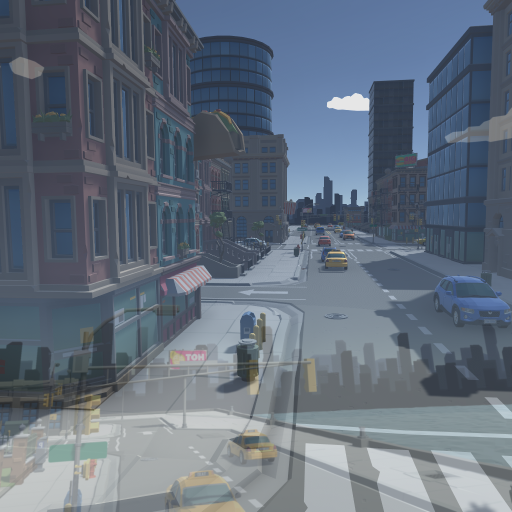}
                            };
                            \begin{scope}[x={(image.south east)},y={(image.north west)}]
                                \node [anchor=south west, fill=white, inner sep=0.1cm] at (0.025,0.035) {$I_2$};
                            \end{scope}
                        \end{tikzpicture}
                    \end{animateinline}
                };
            \end{scope}\end{tikzpicture}
        &
            \begin{tikzpicture}[spy using outlines={circle, 3787CF, magnification=7, size=2.1cm, connect spies,
    every spy in node/.append style={line width=0.04cm}}]
                \node [inner sep=0.0cm] {\includegraphics[width=\itemwidth, interpolate=false, trim={0.0cm 0.0cm 0.0cm 6.0cm}, clip]{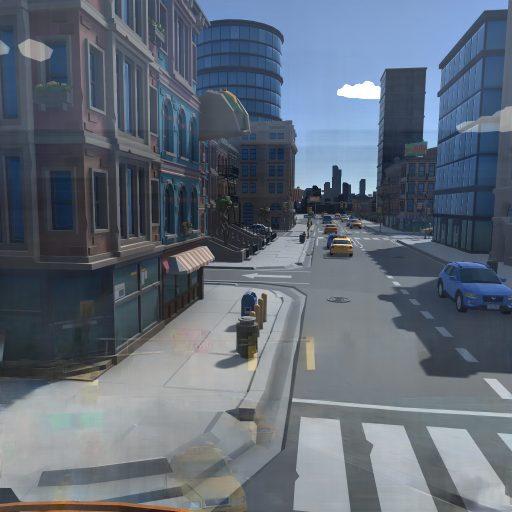}};
                \spy [every spy on node/.append style={line width=0.06cm}, spy connection path={\draw[line width=0.06cm] (tikzspyonnode) -- (tikzspyinnode);}] on (-1.4,-0.65) in node at (0.6,0.0);
            \end{tikzpicture}
        &
            \begin{tikzpicture}[spy using outlines={circle, 3787CF, magnification=7, size=2.1cm, connect spies,
    every spy in node/.append style={line width=0.04cm}}]
                \node [inner sep=0.0cm] {\includegraphics[width=\itemwidth, interpolate=false, trim={0.0cm 0.0cm 0.0cm 6.0cm}, clip]{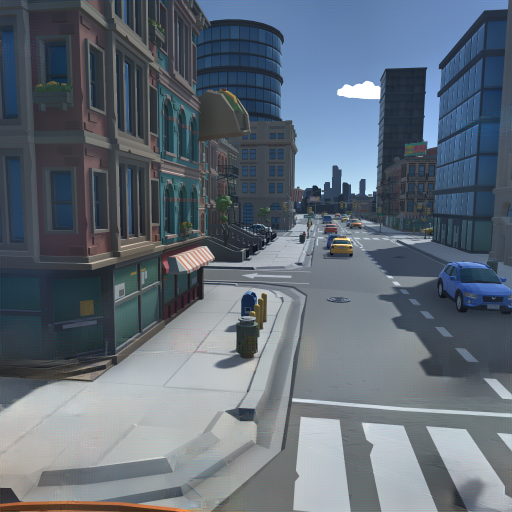}};
                \spy [every spy on node/.append style={line width=0.06cm}, spy connection path={\draw[line width=0.06cm] (tikzspyonnode) -- (tikzspyinnode);}] on (-1.4,-0.65) in node at (0.6,0.0);
            \end{tikzpicture}
        &
            \begin{tikzpicture}[spy using outlines={circle, 3787CF, magnification=7, size=2.1cm, connect spies,
    every spy in node/.append style={line width=0.04cm}}]
                \node [inner sep=0.0cm] {\includegraphics[width=\itemwidth, interpolate=false, trim={0.0cm 0.0cm 0.0cm 6.0cm}, clip]{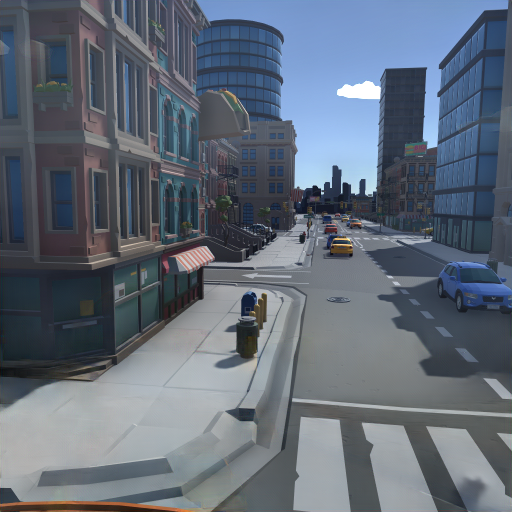}};
                \spy [every spy on node/.append style={line width=0.06cm}, spy connection path={\draw[line width=0.06cm] (tikzspyonnode) -- (tikzspyinnode);}] on (-1.4,-0.65) in node at (0.6,0.0);
            \end{tikzpicture}
        \\
            \footnotesize (a) Input
        &
            \footnotesize (b) $\mathcal{L}_1$
        &
            \footnotesize (c) $\mathcal{L}_F$
        &
            \footnotesize (d) $\mathcal{L}_\text{LPIPS}$
        \\
    \end{tabular}\vspace{-0.2cm}
	\caption{Training with $\ell_1$ distance led to low-frequency artifacts (b), and using squared distance between VGG features led to checkerboard artifacts (c). We hence train our synthesis model using LPIPS, which produces good results (d).}\vspace{-0.4cm}
	\label{fig:loss}
\end{figure*}

\subsection{Reflection-Invariant Optical Flow}
\label{sec:invariant_flow}

Most learning-based optical flow models assume that each pixel has a single motion and train on datasets where this assumption holds~\cite{Butler_ECCV_2012, Dosovitskiy_ICCV_2015}. However, in the presence of reflections, each pixel can have two valid motions: that of the transmission and that of the reflection. Applying learned flow models trained on existing datasets to images containing reflections produces motion estimates that are a compromise between the two true underlying motions, causing them to work poorly for our dereflection task.
We hence train a reflection-invariant flow estimation network using our own synthetic dataset which we introduce in Section~\ref{sec:dataset}. We do so by adopting the architecture of PWC-Net~\cite{Sun_CVPR_2018} and supervising it for $1.5 \cdot 10^6$ iterations with $8$ samples per batch and a learning rate of $10^{-4}$ using TensorFlow's default Adam~\cite{Kingma_ARXIV_2014} optimizer on our new synthetic dataset.

Thanks to our new dataset, our flow model is largely invariant to reflections. In comparison, a model supervised on a reflection-free version of our dataset is subject to a significant drop in its flow prediction accuracy once reflections are introduced (Section~\ref{sec:exp-flow}). This reflection-invariant flow estimate is critical to make our dereflection approach work and an ablation of our pipeline with a regular optical flow network fails to produce convincing results (Section~\ref{sec:exp-syn}).

\subsection{Dual-View Transmission Synthesis}
\label{sec:synthesis}

Given the first view $I_1$ and the aligned second view $I_{2 \shortto 1}$, we utilize a neural network to synthesize the desired transmissive layer $T_1$ of $I_1$. In doing so, the aligned view $I_{2 \shortto 1}$ provides important cues which allow the synthesis network to produce high-quality results despite the presence of significant reflections. Because our optical flow network produces motion estimates that are invariant to reflections, transmissive image content in these warped images is aligned but reflective content is not aligned as long as there is motion parallax between them. This reduces the burden on the synthesis model, as even a pixel-wise minimum of two images should produce good results, as demonstrated in~\cite{Szeliski_CVPR_2000}.

We use a GridNet~\cite{Fourure_BMVC_2017} with the modifications from Niklaus~\etal~\cite{Niklaus_CVPR_2018} for our synthesis network, using five rows and four columns where the first two columns perform downsampling and the last two columns perform upsampling. GridNets are a generalization of U-Nets~\cite{Ronneberger_ARXIV_2015}, which are often used for image synthesis tasks. In essence, GridNets allow information within the network to be processed along multiple streams at different resolutions, which enables them to learn how to combine features across different scales.

We supervise this synthesis model on our dual-view dataset, which we describe in Section~\ref{sec:dataset}. Instead of directly using the ground truth optical flow to warp $I_2$ towards $I_1$, we use the prediction of our reflection-invariant optical flow network. This forces the trained synthesis model to be more robust with respect to misaligned transmissions that may be introduced by erroneous optical flow estimates.

\begin{figure*}\centering
    \setlength{\tabcolsep}{0.05cm}
    \setlength{\itemwidth}{4.26cm}
    \hspace*{-\tabcolsep}\begin{tabular}{cccc}
            \includegraphics[width=\itemwidth]{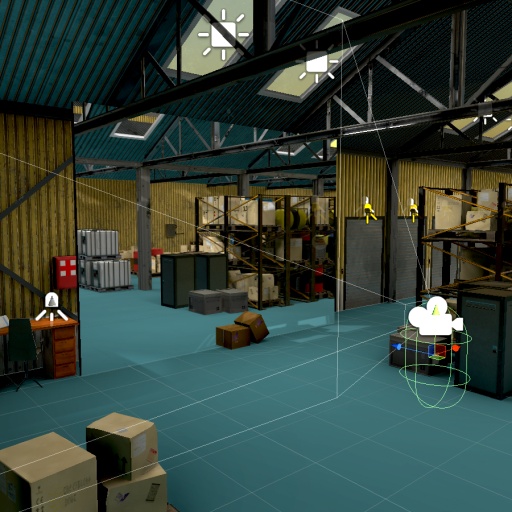}
        &
            \begin{tikzpicture}[scale=4.26]
                \begin{scope}
                    \clip (0,1) -- (0,0.5) -- (1,0.5) -- (1,1) -- cycle;
                    \fill [fill overzoom image=graphics/dataset/render-8-lossy] (0,0) rectangle (1,1);
                    \node [anchor=south west, fill=white, inner sep=0.1cm] at (0.025,0.525) {$I_1$};
                \end{scope}
                \begin{scope}
                    \clip (1,0) -- (1,0.5) -- (0,0.5) -- (0,0) -- cycle;
                    \fill [fill overzoom image=graphics/dataset/render-7-lossy] (0,0) rectangle (1,1);
                    \node [anchor=south west, fill=white, inner sep=0.1cm] at (0.025,0.025) {$I_2$};
                \end{scope}
                \draw [white, line width=0.03cm] (0,0.5) -- (1,0.5);
                \node [inner sep=0.0cm] at (0.5,0.5) {
                    \begin{animateinline}[autoplay, palindrome, final, nomouse, method=widget, poster=none]{0.5}
                        \begin{tikzpicture}
                            \node [anchor=south west, inner sep=0.0cm] (image) at (0,0) {
                                \includegraphics[width=\itemwidth]{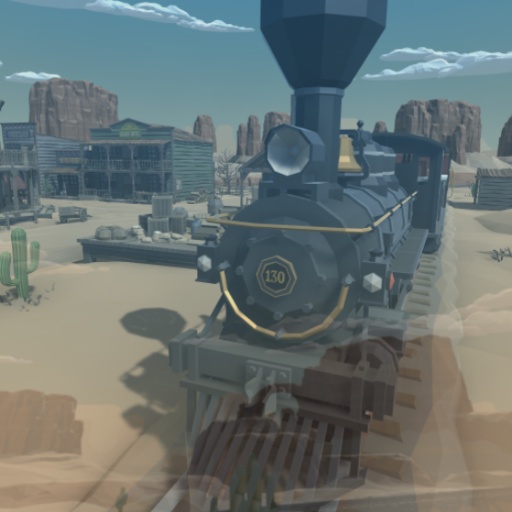}
                            };
                            \begin{scope}[x={(image.south east)},y={(image.north west)}]
                                \node [anchor=south west, fill=white, inner sep=0.1cm] at (0.025,0.025) {$I_1$};
                            \end{scope}
                        \end{tikzpicture}
                        \newframe
                        \begin{tikzpicture}
                            \node [anchor=south west, inner sep=0.0cm] (image) at (0,0) {
                                \includegraphics[width=\itemwidth]{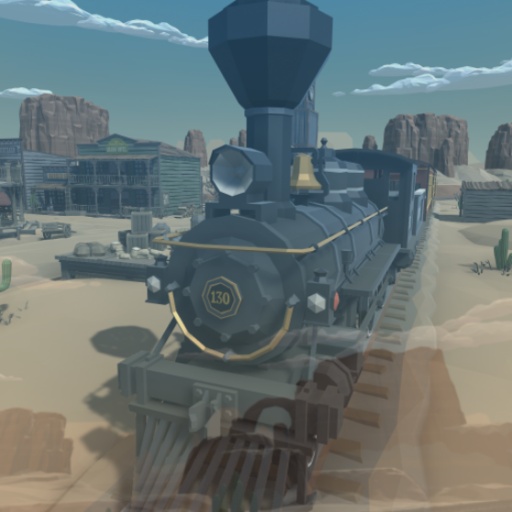}
                            };
                            \begin{scope}[x={(image.south east)},y={(image.north west)}]
                                \node [anchor=south west, fill=white, inner sep=0.1cm] at (0.025,0.025) {$I_2$};
                            \end{scope}
                        \end{tikzpicture}
                    \end{animateinline}
                };
            \end{tikzpicture}
        &
            \begin{tikzpicture}[scale=4.26]
                \begin{scope}
                    \clip (0,1) -- (0,0.5) -- (1,0.5) -- (1,1) -- cycle;
                    \fill [fill overzoom image=graphics/dataset/naive-5-lossy] (0,0) rectangle (1,1);
                    \node [anchor=south west, fill=white, inner sep=0.1cm] at (0.025,0.525) {$I_1$};
                \end{scope}
                \begin{scope}
                    \clip (1,0) -- (1,0.5) -- (0,0.5) -- (0,0) -- cycle;
                    \fill [fill overzoom image=graphics/dataset/naive-4-lossy] (0,0) rectangle (1,1);
                    \node [anchor=south west, fill=white, inner sep=0.1cm] at (0.025,0.025) {$I_2$};
                \end{scope}
                \draw [white, line width=0.03cm] (0,0.5) -- (1,0.5);
                \node [inner sep=0.0cm] at (0.5,0.5) {
                    \begin{animateinline}[autoplay, palindrome, final, nomouse, method=widget, poster=none]{0.5}
                        \begin{tikzpicture}
                            \node [anchor=south west, inner sep=0.0cm] (image) at (0,0) {
                                \includegraphics[width=\itemwidth]{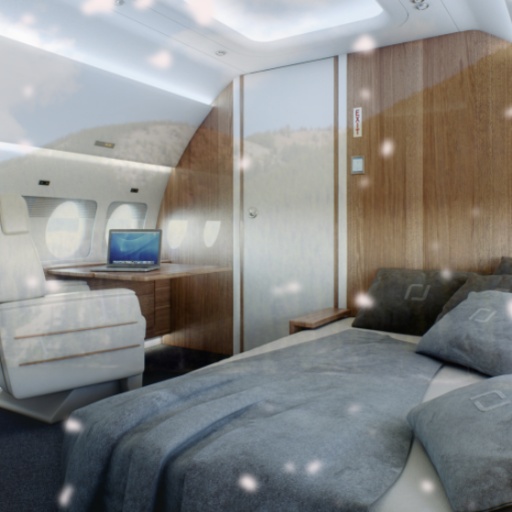}
                            };
                            \begin{scope}[x={(image.south east)},y={(image.north west)}]
                                \node [anchor=south west, fill=white, inner sep=0.1cm] at (0.025,0.025) {$I_{1\vphantom{2 \shortto 1}}$};
                            \end{scope}
                        \end{tikzpicture}
                        \newframe
                        \begin{tikzpicture}
                            \node [anchor=south west, inner sep=0.0cm] (image) at (0,0) {
                                \includegraphics[width=\itemwidth]{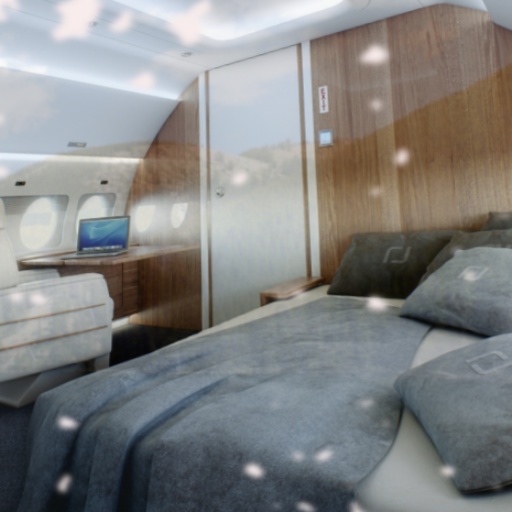}
                            };
                            \begin{scope}[x={(image.south east)},y={(image.north west)}]
                                \node [anchor=south west, fill=white, inner sep=0.1cm] at (0.025,0.025) {$I_2$};
                            \end{scope}
                        \end{tikzpicture}
                    \end{animateinline}
                };
            \end{tikzpicture}
        &
            \begin{tikzpicture}[scale=4.26]
                \begin{scope}
                    \clip (0,1) -- (0,0.5) -- (1,0.5) -- (1,1) -- cycle;
                    \fill [fill overzoom image=graphics/dataset/homography-8-lossy] (0,0) rectangle (1,1);
                    \node [anchor=south west, fill=white, inner sep=0.1cm] at (0.025,0.525) {$I_1$};
                \end{scope}
                \begin{scope}
                    \clip (1,0) -- (1,0.5) -- (0,0.5) -- (0,0) -- cycle;
                    \fill [fill overzoom image=graphics/dataset/homography-7-lossy] (0,0) rectangle (1,1);
                    \node [anchor=south west, fill=white, inner sep=0.1cm] at (0.025,0.025) {$I_2$};
                \end{scope}
                \draw [white, line width=0.03cm] (0,0.5) -- (1,0.5);
                \node [inner sep=0.0cm] at (0.5,0.5) {
                    \begin{animateinline}[autoplay, palindrome, final, nomouse, method=widget, poster=none]{0.5}
                        \begin{tikzpicture}
                            \node [anchor=south west, inner sep=0.0cm] (image) at (0,0) {
                                \includegraphics[width=\itemwidth]{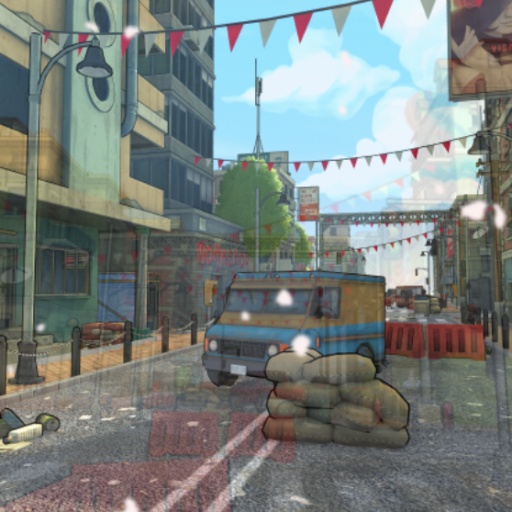}
                            };
                            \begin{scope}[x={(image.south east)},y={(image.north west)}]
                                \node [anchor=south west, fill=white, inner sep=0.1cm] at (0.025,0.025) {$I_1$};
                            \end{scope}
                        \end{tikzpicture}
                        \newframe
                        \begin{tikzpicture}
                            \node [anchor=south west, inner sep=0.0cm] (image) at (0,0) {
                                \includegraphics[width=\itemwidth]{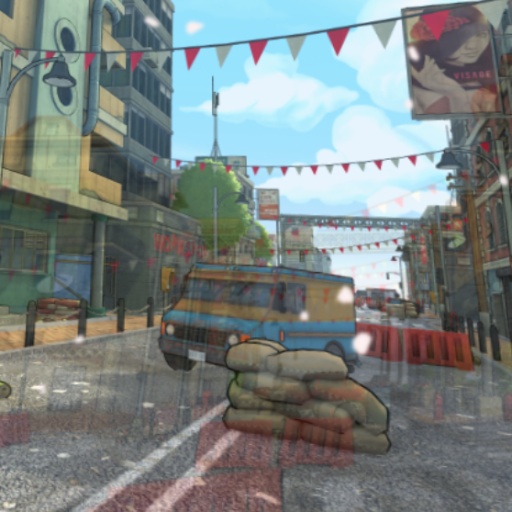}
                            };
                            \begin{scope}[x={(image.south east)},y={(image.north west)}]
                                \node [anchor=south west, fill=white, inner sep=0.1cm] at (0.025,0.025) {$I_2$};
                            \end{scope}
                        \end{tikzpicture}
                    \end{animateinline}
                };
            \end{tikzpicture}
        \\
            \footnotesize (a) Unity Editor
        &
            \footnotesize (b) Rendered Views
        &
            \footnotesize (c) Warped Images
        &
            \footnotesize (d) Warped Renders
        \\        
    \end{tabular}\vspace{-0.2cm}
	\caption{Our training dataset consists of three different types of images: 60\% are fully-rendered images generated using the Unity engine (a) and consist of scenes with complex geometry (b), 30\% are real images that lack ground-truth geometry and have instead been warped using random homographies to generate the second view (c), and 10\% are warped rendered images to make sure that the model does not ``cheat'' (d). Note that because (b) is fully rendered, its reflective layer originates from the same domain as the transmissive layer (both are mountains), while the two layers in (c) may have different sources.}\vspace{-0.4cm}
	\label{fig:dataset}
\end{figure*}
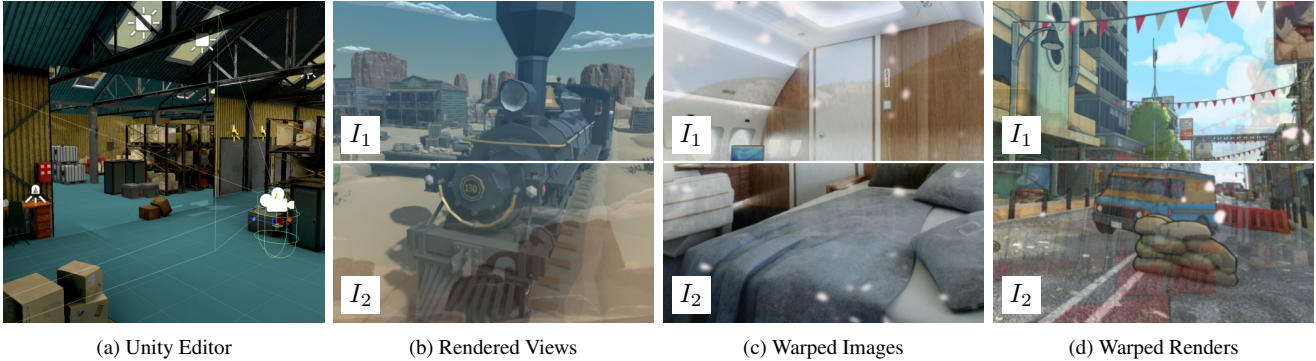

We analyzed several possible loss functions to supervise our synthesis model. The simplest of which is the $\ell_1$ distance between the predicted transmission layer and ground truth. However, a synthesis model supervised with just $\mathcal{L}_{1}$ is prone to low-frequency artifacts as shown in Figure~\ref{fig:loss}(b). We additionally explored a loss based on the squared distance between VGG features~\cite{Johnson_ECCV_2016}, which some recent dereflection algorithms have used successfully~\cite{Xuzhang_CVPR_2018}. However, we noticed subtle checkerboard artifacts when supervising our synthesis model on this $\mathcal{L}_F$ as shown in Figure~\ref{fig:loss}(c) (even when using bilinear upsampling instead of transposed convolutions~\cite{Odena_OTHER_2016}). We thus used the LPIPS metric~\cite{Rizhang_CVPR_2018}, which linearly weights feature activations using a channel-wise vector $w$ as
\begin{equation}
    \mathcal{L}_\text{LPIPS} = \sum_\ell \left\|  w^{\ell} \odot \left( \Phi^{\ell} \left( T_1^{\mathit{pred}} \right) - \Phi^{\ell} \left(T_1^{\mathit{gt}} \right) \right) \right\| _2^2.
\end{equation}
Specifically, we use version ``0.1'' of this metric, using AlexNet~\cite{Krizhevsky_NIPS_2012} to compute feature activations, and where the weights $w$ have been linearly calibrated to minimize the perceptual difference in accordance with a user study~\cite{Rizhang_CVPR_2018}. Our synthesis model trained using $\mathcal{L}_\text{LPIPS}$ is able to produce pleasant results that are not subject to checkerboard artifacts, as shown in Figure~\ref{fig:loss}(d).
This perceptual loss serves a similar purpose as adversarial losses, which have also been an effective mean for the task of reflection removal~\cite{Xuzhang_CVPR_2018}.

We train our proposed dual-view transmission synthesis model using TensorFlow's default Adam~\cite{Kingma_ARXIV_2014} optimizer with a learning rate of $5 \cdot 10^{-5}$, which took a total of $1.5$ million iterations with $4$ samples per batch to fully converge.

\subsection{Dual-View Training Data}
\label{sec:dataset}

Existing learning-based methods for dereflection combine pairs of images to synthesize training data~\cite{Fan_ICCV_2017, Xuzhang_CVPR_2018}. 
This approach works well for monocular approaches, but it does not generalize to our dual-view approach. After all, whatever reflection we add to a stereo pair should be geometrically consistent across the two views which requires difficult-to-acquire depth maps. Furthermore, training our reflection-invariant flow network requires ground truth optical flow between the transmissive layers of the two views. However, acquiring ground truth flow is a challenging problem with previous work having exploited hidden fluorescent textures, computer graphics, and high frame-rate videos~\cite{Baker_IJCV_2011, Butler_ECCV_2012, Janai_CVPR_2017}.

For these reasons, we rely on computer graphics to synthesize our training data. We acquired $20$ virtual environments from professional artists, $17$ of which are used for training and $3$ of which are used for evaluation. These environments vary greatly, and include indoor scenes, cityscapes, and naturalistic scenes. We render them with Unity, which allowed us to collect arbitrary views together with a ground-truth inter-frame optical flow. Views are generated by pre-recording camera paths through the scene, from which we sample camera locations for $I_1$. We generate $I_2$ by randomly shifting the position of $I_1$ by up to $0.5$ meters and randomly rotating the camera by up to 10 degrees. To model reflections, we create a translucent mirror that is placed in front of the two cameras. We uniformly sample the mirror's alpha blending factor $\alpha \sim \mathcal{U}(0.6, 0.9)$, and apply a Gaussian blur with a random $\sigma \sim \mathcal{U}(0.0, 0.1)$ to the reflective image to mimic depth of field. We then alpha-blend the transmissive and reflective images to get the rendered output for $I_1$ and $I_2$.

Training only on synthetic data may result in poor performance on real-world data, due to a significant gap between the two domains~\cite{Mayer_IJCV_2018}. To address this, we augment our synthetic data with additional training data that has been generated using real-world images. We first randomly sample two images and blend them to get the input for one view, and apply two homography transforms to the two images independently to synthesize the image in the other view. This basically assumes that the transmissive and reflective layers are on independent planes. Although this over-simplifies the geometry of the real world compared with our fully-rendered data, it helps the network to better fit to the statistics of real-world images. We collected $7000$ images with a Creative Commons license for this purpose and manually selected those with pleasant visual aesthetics, which yielded a subset of $1000$ images in total. As shown Figure~\ref{fig:dataset}(c), this data is closer to real world imagery but it lacks real motion parallax. Warping image $I_2$ to image $I_1$ according to the transmission flow is hence free from disocclusions. This is not the only unrealistic aspect of this approach though, since reflections may not originate form the same scene like as in the picture of a hotel room that exhibits reflections of a mountain.

To make sure that our model does not ``cheat'' by identifying which images are real and taking advantage of our simple proxy geometry, we also applied the same homography-based image formation model that was used for our real-world data to our rendered data, as shown in Figure~\ref{fig:dataset}(d).

Lastly, many reflections in the real world stem from light sources which yield saturated bright spots in the image. To model this, we augment the reflective layer with a mask of bright spots obtained from binarized fractal noise: we compute the fractal noise from Perlin noise at 4 octaves with a persistence uniformly drawn from $\rho \sim \mathcal{U}(0.3, 1.0)$ before binarizing the mask based on a threshold of $1$. To avoid unnatural discontinuities, we further apply a Gaussian blur with $\sigma \sim \mathcal{U}(1, 5)$ to this binary mask. Examples of such saturated bright spots are shown in Figure~\ref{fig:dataset}(c) and \ref{fig:dataset}(d).

When using this training dataset, we randomly sample 60\% of the batches from our rendered data, 30\% from our warped images, and 10\% from our warped renderings.

%% file: experiments.tex
We evaluate on rendered and real-world images.

\vspace{0.05in}
\noindent\textbf{Rendered test set:} To build a rendered test set, we used $3$ virtual worlds that are not used in training and rendered $60$ different samples. We also recorded the corresponding ground truth transmission image without reflection and the ground truth optical flow between the transmission layers.

\begin{figure}\centering
    \setlength{\tabcolsep}{0.05cm}
    \setlength{\itemwidth}{4.11cm}
    \hspace*{-\tabcolsep}\begin{tabular}{cc}
            \includegraphics[width=\itemwidth, trim={0.0cm 10.9cm 0.0cm 0.0cm}, clip]{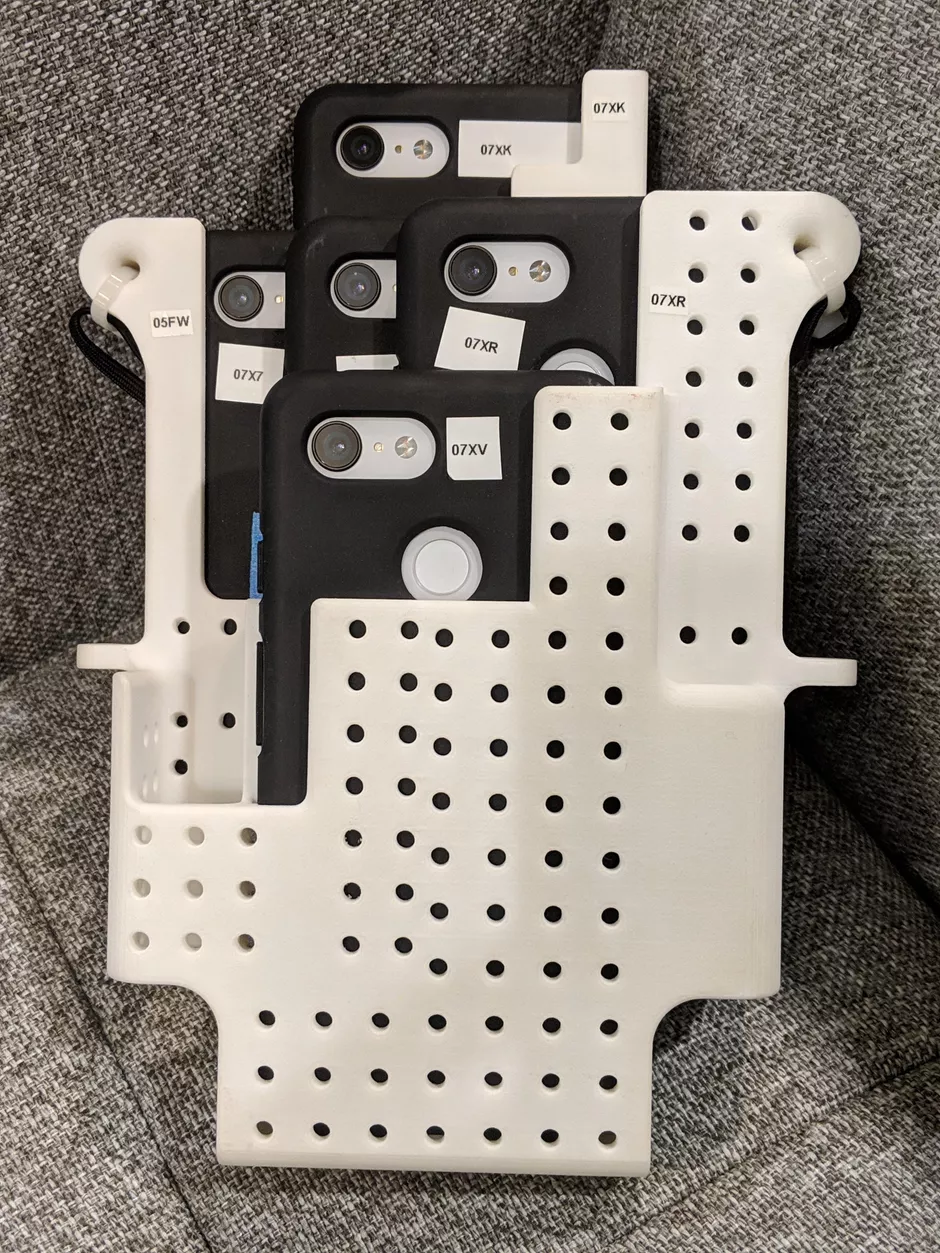}
        &
            \includegraphics[width=\itemwidth, trim={0.0cm 17.6cm 0.0cm 17.6cm}, clip]{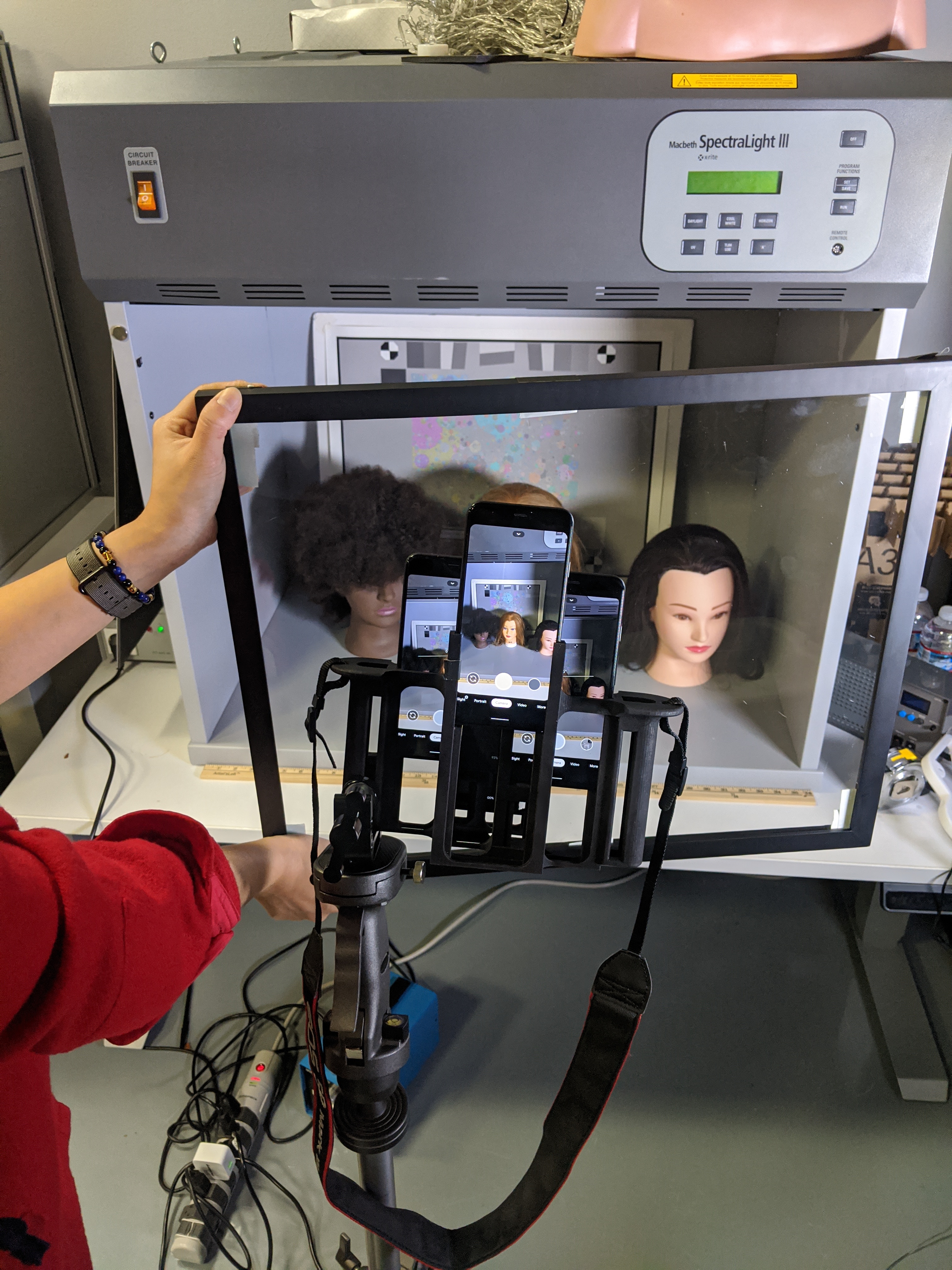}
        \\
            \footnotesize (a) Camera
        &
            \footnotesize (b) Setup
        \\
    \end{tabular}\vspace{-0.2cm}
	\caption{A picture of our custom-built camera rig consisting of five synchronized Google Pixel phones (a) as well as a schematic reenactment of the data capturing setup (b).}\vspace{-0.2cm}
	\label{fig:camsetup}
\end{figure}

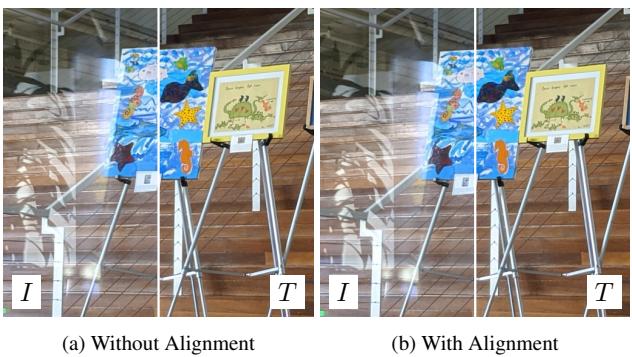
\begin{figure}\centering
    \setlength{\tabcolsep}{0.05cm}
    \setlength{\itemwidth}{4.11cm}
    \hspace*{-\tabcolsep}\begin{tabular}{cc}
            \begin{tikzpicture}[scale=4.11]
                \begin{scope}
                    \clip (0,1) -- (0.5,1) -- (0.5,0) -- (0,0) -- cycle;
                    \fill [fill overzoom image=graphics/frankencam/before-input-crop] (0,0) rectangle (1,1);
                    \node [anchor=south west, fill=white, inner sep=0.1cm] at (0.025,0.025) {$I$};
                \end{scope}
                \begin{scope}
                    \clip (1,1) -- (0.5,1) -- (0.5,0) -- (1,0) -- cycle;
                    \fill [fill overzoom image=graphics/frankencam/before-gt-crop] (0,0) rectangle (1,1);
                    \node [anchor=south east, fill=white, inner sep=0.1cm] at (0.975,0.025) {$T$};
                \end{scope}
                \draw [white, line width=0.03cm] (0.5,0) -- (0.5,1);
                \node [inner sep=0.0cm] at (0.5,0.5) {
                    \begin{animateinline}[autoplay, palindrome, final, nomouse, method=widget, poster=none]{1}
                        \begin{tikzpicture}
                            \node [anchor=south west, inner sep=0.0cm] (image) at (0,0) {
                                \includegraphics[width=\itemwidth]{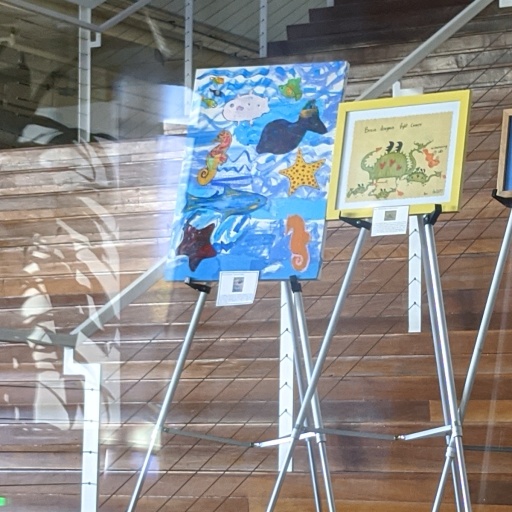}
                            };
                            \begin{scope}[x={(image.south east)},y={(image.north west)}]
                                \node [anchor=south west, fill=white, inner sep=0.1cm] at (0.025,0.025) {$I$};
                            \end{scope}
                        \end{tikzpicture}
                        \newframe
                        \begin{tikzpicture}
                            \node [anchor=south west, inner sep=0.0cm] (image) at (0,0) {
                                \includegraphics[width=\itemwidth]{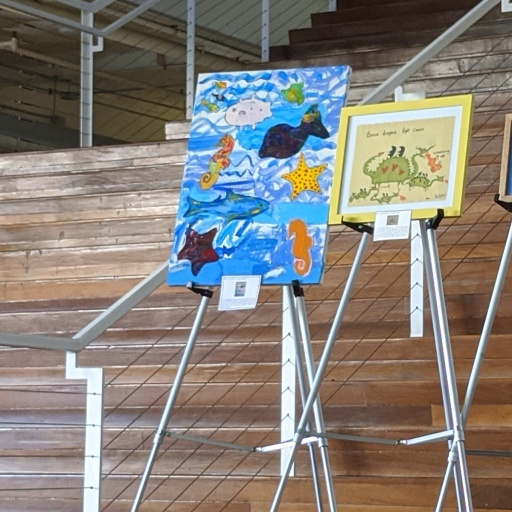}
                            };
                            \begin{scope}[x={(image.south east)},y={(image.north west)}]
                                \node [anchor=south west, fill=white, inner sep=0.1cm] at (0.025,0.025) {$T$};
                            \end{scope}
                        \end{tikzpicture}
                    \end{animateinline}
                };
            \end{tikzpicture}
        &
            \begin{tikzpicture}[scale=4.11]
                \begin{scope}
                    \clip (0,1) -- (0.5,1) -- (0.5,0) -- (0,0) -- cycle;
                    \fill [fill overzoom image=graphics/frankencam/after-input-crop] (0,0) rectangle (1,1);
                    \node [anchor=south west, fill=white, inner sep=0.1cm] at (0.025,0.025) {$I$};
                \end{scope}
                \begin{scope}
                    \clip (1,1) -- (0.5,1) -- (0.5,0) -- (1,0) -- cycle;
                    \fill [fill overzoom image=graphics/frankencam/after-gt-crop] (0,0) rectangle (1,1);
                    \node [anchor=south east, fill=white, inner sep=0.1cm] at (0.975,0.025) {$T$};
                \end{scope}
                \draw [white, line width=0.03cm] (0.5,0) -- (0.5,1);
                \node [inner sep=0.0cm] at (0.5,0.5) {
                    \begin{animateinline}[autoplay, palindrome, final, nomouse, method=widget, poster=none]{1}
                        \begin{tikzpicture}
                            \node [anchor=south west, inner sep=0.0cm] (image) at (0,0) {
                                \includegraphics[width=\itemwidth]{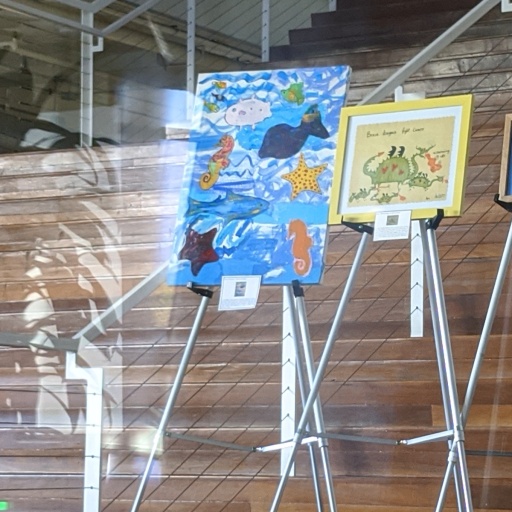}
                            };
                            \begin{scope}[x={(image.south east)},y={(image.north west)}]
                                \node [anchor=south west, fill=white, inner sep=0.1cm] at (0.025,0.025) {$I$};
                            \end{scope}
                        \end{tikzpicture}
                        \newframe
                        \begin{tikzpicture}
                            \node [anchor=south west, inner sep=0.0cm] (image) at (0,0) {
                                \includegraphics[width=\itemwidth]{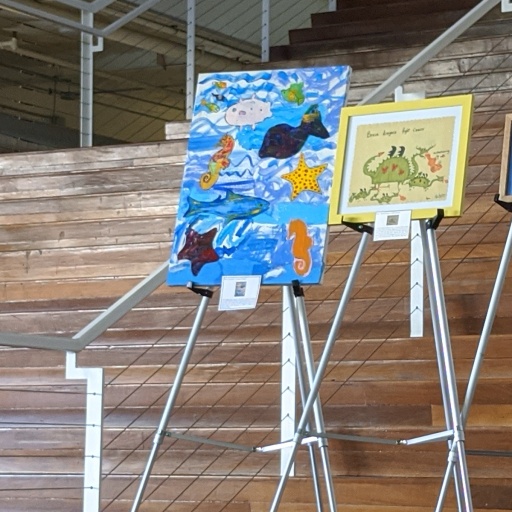}
                            };
                            \begin{scope}[x={(image.south east)},y={(image.north west)}]
                                \node [anchor=south west, fill=white, inner sep=0.1cm] at (0.025,0.025) {$T$};
                            \end{scope}
                        \end{tikzpicture}
                    \end{animateinline}
                };
            \end{tikzpicture}
        \\
            \footnotesize (a) Without Alignment
        &
            \footnotesize (b) With Alignment
        \\
    \end{tabular}\vspace{-0.2cm}
	\caption{The images in our dataset and their respective transmissions are misaligned due to refraction (a), as can be seen at the stairs. We align them to account for this (b).}\vspace{-0.4cm}
	\label{fig:camalign}
\end{figure}

\vspace{0.05in}
\noindent\textbf{Real-world test set:} To build a real-world test set, we use a camera rig of five phones as shown in Figure~\ref{fig:camsetup} and synchronize them using~\cite{Ansari_OTHER_2019}. To test that our approach works for different stereo configurations, we always use the center camera as the reference view and one of the other four cameras as the second view. For each of the $20$ scenes we captured, we obtained the transmission and between $2$ and $4$ sets of images with reflections by placing different types of glass in front of the camera. As discussed in~\cite{Xue_TOG_2015}, the transmission shifts between the image capturing with the glass and without the glass due to refractions unless the glass is infinitely thin. Therefore, we register the image captured through glass to the ground truth transmission (image captured without glass) using an affine transform calculated by~\cite{Evangelidis_PAMI_2008}. An example of this alignment is shown in Figure~\ref{fig:camalign}.

\subsection{Reflection-Invariant Optical Flow}
\label{sec:exp-flow}

\noindent\textbf{Metrics:} Following optical flow literature~\cite{Baker_IJCV_2011}, we use two metrics to evaluate flow accuracy: 1) the end-point error (EPE) between the estimated flow and the true flow, and 2) the absolute difference (ABS) between the first frame and the second frame warped to the first frame using the estimated flow. For the ABS metric, as we only calculate the motion of the transmission layer, we only warp the ground truth transmission layer without reflection even though the motion was estimated from the input images with reflection. We also mask out the occluded pixels based on the true transmission optical flow when calculating the ABS metric.

\begin{figure}\centering
	\setlength{\tabcolsep}{0.0cm}
	\renewcommand{\arraystretch}{1.2}
	\newcommand{\quantTit}[1]{\multicolumn{4}{c}{\scriptsize #1}}
	\newcommand{\quantSec}[1]{\scriptsize #1}
	\newcommand{\quantInd}[1]{\tiny #1}
	\newcommand{\quantVal}[1]{\scalebox{0.83}[1.0]{$ #1 $}}
	\newcommand{\quantBes}[1]{\scalebox{0.83}[1.0]{$\uline{ #1 }$}}
	\footnotesize
	\begin{tabularx}{\columnwidth}{@{\hspace{0.1cm}} X P{0.92cm} @{\hspace{-0.2cm}} P{0.92cm} @{\hspace{-0.2cm}} P{0.92cm} @{\hspace{-0.2cm}} P{0.92cm} P{0.92cm} @{\hspace{-0.2cm}} P{0.92cm} @{\hspace{-0.2cm}} P{0.92cm} @{\hspace{-0.2cm}} P{0.92cm}}
		\toprule
			& \quantTit{rendered test w/o refl.} & \quantTit{rendered test w/ refl.}
		\\ \cmidrule(l{2pt}r{2pt}){2-5} \cmidrule(l{2pt}r{2pt}){6-9}
			& \quantSec{EPE} \linebreak \quantInd{mean} & \quantSec{EPE} \linebreak \quantInd{median} & \quantSec{ABS} \linebreak \quantInd{mean} & \quantSec{ABS} \linebreak \quantInd{median} & \quantSec{EPE} \linebreak \quantInd{mean} & \quantSec{EPE} \linebreak \quantInd{median} & \quantSec{ABS} \linebreak \quantInd{mean} & \quantSec{ABS} \linebreak \quantInd{median}
		\\ \midrule
Zeros & \quantVal{24.90} & \quantVal{22.88} & \quantVal{24.54} & \quantVal{24.00} & \quantVal{24.90} & \quantVal{22.88} & \quantVal{24.54} & \quantVal{24.00}
\\
Oracle & \quantVal{0.0} & \quantVal{0.0} & \quantVal{3.13} & \quantVal{2.88} & \quantVal{0.0} & \quantVal{0.0} & \quantVal{3.13} & \quantVal{2.88}
\\ \midrule
Train w/o refl. & \quantBes{1.14} & \quantBes{0.84} & \quantBes{4.02} & \quantBes{3.56} & \quantVal{4.52} & \quantVal{2.67} & \quantVal{6.10} & \quantVal{5.56}
\\
Train w/ refl. & \quantVal{1.53} & \quantVal{1.05} & \quantVal{4.23} & \quantVal{3.67} & \quantBes{2.39} & \quantBes{1.26} & \quantBes{4.68} & \quantBes{4.06}
		\\ \bottomrule
	\end{tabularx}\vspace{-0.1cm}
	\captionof{table}{Flow accuracy on our rendered test set. We trained two versions of our flow network, one using our rendered test set w/ reflections and one w/o reflections. We also report the accuracy of zero and ground truth motion as bounds.}\vspace{-0.2cm}
	\label{tbl:flowtable}
\end{figure}

\begin{figure}\centering
    \setlength{\tabcolsep}{0.05cm}
    \setlength{\itemwidth}{4.11cm}
    \hspace*{-\tabcolsep}\begin{tabular}{cc}
            \begin{tikzpicture}
                \definecolor{arrowcolor}{RGB}{238,127,14}
                \node [anchor=south west, inner sep=0.0cm] (image) at (0,0) {
                    \includegraphics[width=\itemwidth, trim={1.0cm 0.0cm 0.0cm 0.0cm}, clip]{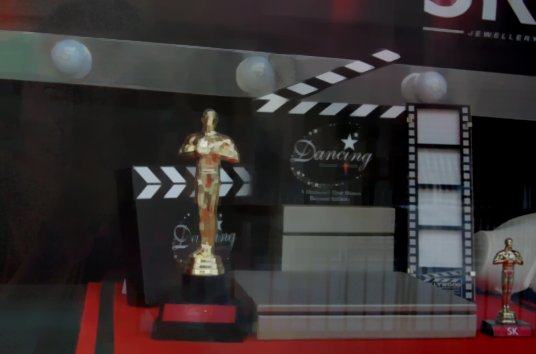}
                };
                \begin{scope}[x={(image.south east)},y={(image.north west)}]
                    \draw [double arrow=0.2cm with white and arrowcolor] (0.93,0.5) -- (0.73,0.25);
                \end{scope}
            \end{tikzpicture}
        &
            \begin{tikzpicture}
                \definecolor{arrowcolor}{RGB}{97,157,71}
                \node [anchor=south west, inner sep=0.0cm] (image) at (0,0) {
                    \includegraphics[width=\itemwidth, trim={1.0cm 0.0cm 0.0cm 0.0cm}, clip]{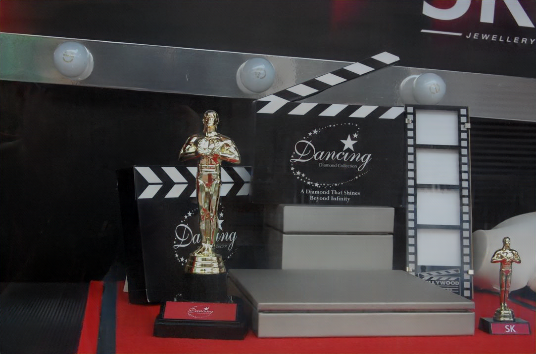}
                };
                \begin{scope}[x={(image.south east)},y={(image.north west)}]
                    \draw [double arrow=0.2cm with white and arrowcolor] (0.93,0.5) -- (0.73,0.25);
                \end{scope}
            \end{tikzpicture}
        \\
            \begin{tikzpicture}
                \definecolor{arrowcolor}{RGB}{238,127,14}
                \node [anchor=south west, inner sep=0.0cm] (image) at (0,0) {
                    \includegraphics[width=\itemwidth, trim={0.4cm 0.0cm 0.6cm 0.0cm}, clip]{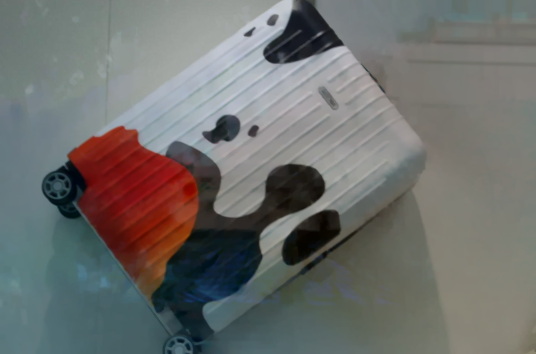}
                };
                \begin{scope}[x={(image.south east)},y={(image.north west)}]
                    \draw [double arrow=0.2cm with white and arrowcolor] (0.6,0.17) -- (0.31,0.2);
                    \draw [double arrow=0.2cm with white and arrowcolor] (0.88,0.25) -- (0.85,0.65);
                \end{scope}
            \end{tikzpicture}
        &
            \begin{tikzpicture}
                \definecolor{arrowcolor}{RGB}{97,157,71}
                \node [anchor=south west, inner sep=0.0cm] (image) at (0,0) {
                    \includegraphics[width=\itemwidth, trim={0.4cm 0.0cm 0.6cm 0.0cm}, clip]{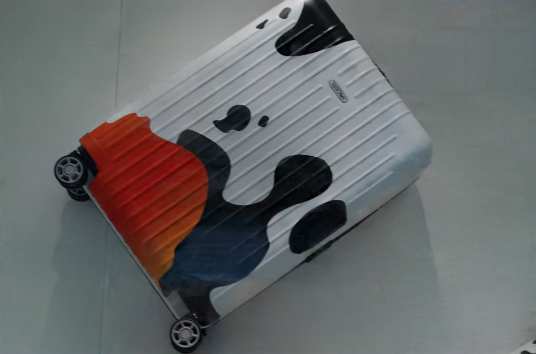}
                };
                \begin{scope}[x={(image.south east)},y={(image.north west)}]
                    \draw [double arrow=0.2cm with white and arrowcolor] (0.6,0.17) -- (0.31,0.2);
                    \draw [double arrow=0.2cm with white and arrowcolor] (0.88,0.25) -- (0.85,0.65);
                \end{scope}
            \end{tikzpicture}
        \\
            \footnotesize (a) Robust Flow~\cite{Yang_CVPR_2016}
        &
            \footnotesize (b) Ours
        \\
    \end{tabular}\vspace{-0.2cm}
	\caption{Comparisons with \cite{Yang_CVPR_2016}, a robust optical flow via classic energy minimization, on examples from their paper.}\vspace{-0.4cm}
	\label{fig:robflow}
\end{figure}

\vspace{0.05in}
\noindent\textbf{Results:} Table~\ref{tbl:flowtable} shows the quantitative results. To better understand the scale of EPE and ABS, we also report these metrics for zero flow (all pixels are static) and ground truth transmission flow (``Oracle''). Note that because of lighting changes between left and right views, the ABS error of the ground truth flow is not zero. When evaluating on input with reflection, the flow network trained with reflection is more robust than the one trained without reflection, with 47\% less mean EPE error and 23\% less mean ABS error. We analyze the effect of this difference in the context of our reflection removal pipeline in the ablation study in Section~\ref{sec:exp-syn}.

\begin{figure}\centering
	\setlength{\tabcolsep}{0.0cm}
	\renewcommand{\arraystretch}{1.2}
	\newcommand{\quantSec}[1]{\scriptsize #1}
	\newcommand{\quantInd}[1]{\tiny #1}
	\newcommand{\quantVal}[1]{\scalebox{0.78}[1.0]{$ #1 $}}
	\newcommand{\quantBes}[1]{\scalebox{0.78}[1.0]{$\uline{ #1 }$}}
	\footnotesize
	\begin{tabularx}{\columnwidth}{@{\hspace{0.1cm}} X P{1.1cm} P{0.3cm} P{0.75cm} P{0.75cm} P{0.75cm} P{0.4cm} P{0.75cm} P{0.75cm} P{0.75cm}}
		\toprule
			& && \multicolumn{3}{c}{\scriptsize rendered test set} && \multicolumn{3}{c}{\scriptsize real-world test set}
		\\ \cmidrule(l{2pt}r{2pt}){4-6} \cmidrule(l{2pt}r{2pt}){8-10}
			& {\vspace{-0.3cm} \scriptsize images \setlength{\parskip}{-0.09cm}\par used} && \quantSec{PSNR} \par \quantInd{$\uparrow$} & \quantSec{SSIM} \par \quantInd{$\uparrow$} & \quantSec{LPIPS} \par \quantInd{$\downarrow$} && \quantSec{PSNR} \par \quantInd{$\uparrow$} & \quantSec{SSIM} \par \quantInd{$\uparrow$} & \quantSec{LPIPS} \par \quantInd{$\downarrow$}
		\\ \midrule
\scalebox{0.9}[1.0]{Zhang-like} & $1$ && \quantVal{23.92} & \quantVal{0.872} & \quantVal{0.137} && \quantVal{22.39} & \quantVal{0.742} & \quantVal{0.124}
\\
\scalebox{0.9}[1.0]{Mono} & $1$ && \quantVal{26.31} & \quantVal{0.928} & \quantVal{0.068} && \quantVal{22.35} & \quantVal{0.752} & \quantVal{0.110}
\\
\scalebox{0.9}[1.0]{Concat} & $2$ && \quantVal{25.81} & \quantVal{0.927} & \quantVal{0.069} && \quantVal{22.10} & \quantVal{0.752} & \quantVal{0.111}
\\
\scalebox{0.9}[1.0]{Regular Flow} & $2$ && \quantVal{17.61} & \quantVal{0.827} & \quantVal{0.099} && \quantVal{17.73} & \quantVal{0.684} & \quantVal{0.128}
\\
\scalebox{0.9}[1.0]{Ours} & $2$ && \quantBes{26.60} & \quantBes{0.938} & \quantBes{0.058} && \quantBes{22.82} & \quantBes{0.765} & \quantBes{0.104}
		\\ \bottomrule
	\end{tabularx}\vspace{-0.12cm}
	\captionof{table}{Results from our ablation study, showing the importance of GridNet and reflection-invariant optical flow.}\vspace{-0.2cm}
	\label{tbl:ablation}
\end{figure}

\begin{figure}\centering
	\setlength{\tabcolsep}{0.0cm}
	\renewcommand{\arraystretch}{1.2}
	\newcommand{\quantSec}[1]{\scriptsize #1}
	\newcommand{\quantInd}[1]{\tiny #1}
	\newcommand{\quantVal}[1]{\scalebox{0.78}[1.0]{$ #1 $}}
	\newcommand{\quantBes}[1]{\scalebox{0.78}[1.0]{$\uline{ #1 }$}}
	\footnotesize
	\begin{tabularx}{\columnwidth}{@{\hspace{0.1cm}} X P{0.75cm} P{0.75cm} P{0.75cm} P{1.0cm} P{0.75cm} P{0.75cm} P{0.75cm} P{1.0cm}}
		\toprule
		    & \multicolumn{4}{c}{\scriptsize rendered test set} & \multicolumn{4}{c}{\scriptsize real-world test set}
		\\ \cmidrule(l{2pt}r{2pt}){2-5} \cmidrule(l{2pt}r{2pt}){6-9}
			& \multicolumn{3}{c}{\scriptsize quantitative} & \multicolumn{1}{c}{\scriptsize users} & \multicolumn{3}{c}{\scriptsize quantitative} & \multicolumn{1}{c}{\scriptsize users}
		\\ \cmidrule(l{2pt}r{2pt}){2-4} \cmidrule(l{2pt}r{2pt}){5-5} \cmidrule(l{2pt}r{2pt}){6-8} \cmidrule(l{2pt}r{2pt}){9-9}
			& \quantSec{PSNR} \par \quantInd{$\uparrow$} & \quantSec{SSIM} \par \quantInd{$\uparrow$} & \quantSec{LPIPS} \par \quantInd{$\downarrow$} & {\vspace{-0.3cm} \scriptsize prefer \setlength{\parskip}{-0.09cm}\par ours} & \quantSec{PSNR} \par \quantInd{$\uparrow$} & \quantSec{SSIM} \par \quantInd{$\uparrow$} & \quantSec{LPIPS} \par \quantInd{$\downarrow$} & {\vspace{-0.3cm} \scriptsize prefer \setlength{\parskip}{-0.09cm}\par ours}
		\\ \midrule
Input & \quantVal{23.38} & \quantVal{0.887} & \quantVal{0.155} & \quantVal{99\%} & \quantVal{22.25} & \quantVal{0.761} & \quantVal{0.114} & \quantVal{95\%}
\\
Zhang~\etal & \quantVal{22.21} & \quantVal{0.811} & \quantVal{0.217} & \quantVal{99\%} & \quantVal{21.47} & \quantVal{0.725} & \quantVal{0.172} & \quantVal{87\%}
\\
Wen~\etal & \quantVal{22.34} & \quantVal{0.856} & \quantVal{0.185} & \quantVal{100\%} & \quantVal{21.56} & \quantVal{0.744} & \quantVal{0.142} & \quantVal{94\%}
\\
Li \& Brown & \quantVal{22.00} & \quantVal{0.794} & \quantVal{0.243} & \quantVal{100\%} & \quantVal{20.49} & \quantVal{0.671} & \quantVal{0.227} & \quantVal{98\%}
\\
Ours - Mono & \quantVal{26.31} & \quantVal{0.928} & \quantVal{0.068} & \quantVal{94\%} & \quantVal{22.35} & \quantVal{0.752} & \quantVal{0.110} & \quantVal{92\%}
\\
Ours & \quantBes{26.60} & \quantBes{0.938} & \quantBes{0.058} & \quantVal{-} & \quantBes{22.82} & \quantBes{0.765} & \quantBes{0.104} & \quantVal{-}
		\\ \bottomrule
	\end{tabularx}\vspace{-0.1cm}
	\captionof{table}{Quantitative evaluation of the recovered transmission image, together with the results from a user study with responses from 20 participants across 9 rendered test images and 20 real test images. Users were asked to compare our dual-view result to one of five baselines. We report the percentage of times that users preferred our method.}\vspace{-0.2cm}
	\label{tbl:quantstudy}
\end{figure}

\vspace{0.05in}
\noindent\textbf{Related:} Optical flow estimation on layered compound images has previously been studied by Yang~\etal\cite{Yang_CVPR_2016}, who proposed a solution based on classic energy minimization. We were unable to use this technique as a baseline on our benchmark, as the implementation provided by the authors does not allow for arbitrary images to be processed (it requires some external optical flow estimate as input). We hence compare to this technique by instead applying our dereflection pipeline to the example images used by~\cite{Yang_CVPR_2016}. As can be seen in Figure~\ref{fig:robflow}, our proposed approach produces significantly improved reflection removal results.

\subsection{Dual-View Transmission Synthesis} 
\label{sec:exp-syn}

\noindent\textbf{Metrics:} To quantitatively evaluate the quality of reflection removal, we use three evaluation metrics: PSNR, the hand-designed similarity metric SSIM proposed by Wang~\etal~\cite{Wang_TIP_2004}, and the learned similarity metric LPIPS proposed by Zhang~\etal~\cite{Rizhang_CVPR_2018}. Because the transmission coefficient of glass is less than $1.0$, the transmission captured through the glass is dimmer than the image captured without glass. As a result, there is an unknown scaling factor between the estimated transmission and the ground truth. To make our evaluation invariant to this unknown scaling factor, we first scale the estimated transmission by searching for the gain $s$ and bias $b$ that minimize $\|s \cdot T^\text{pred}_1 + b - T^\text{gt}_1\|_2$, before computing the error metrics using the scaled estimate.

\begin{figure}[t!]\centering
    \setlength{\tabcolsep}{0.05cm}
    \setlength{\itemwidth}{4.11cm}
    \hspace*{-\tabcolsep}\begin{tabular}{cc}
            \begin{tikzpicture}[spy using outlines={circle, 3787CF, magnification=3, size=2.5cm, connect spies,
    every spy in node/.append style={line width=0.06cm}}]
                \node [inner sep=0.0cm] {\includegraphics[width=\itemwidth, interpolate=true, trim={0.0cm 0.77cm 0.0cm 0.0cm}, clip]{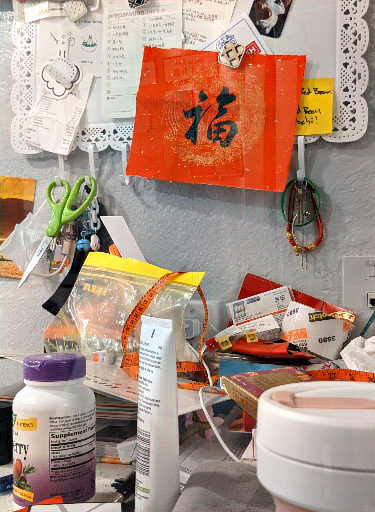}};
                \spy [every spy on node/.append style={line width=0.06cm}, spy connection path={\draw[line width=0.06cm] (tikzspyonnode) -- (tikzspyinnode);}] on (0.5,0.6) in node at (-0.6,-1.25);
                \node [align=center, fill=white, anchor=north west] at (-1.95,2.53) {\footnotesize PSNR: $22.60$};
            \end{tikzpicture}
            &
            \begin{tikzpicture}[spy using outlines={circle, 3787CF, magnification=3, size=2.5cm, connect spies,
    every spy in node/.append style={line width=0.06cm}}]
                \node [inner sep=0.0cm] {\includegraphics[width=\itemwidth, interpolate=true, trim={0.0cm 0.77cm 0.0cm 0.0cm}, clip]{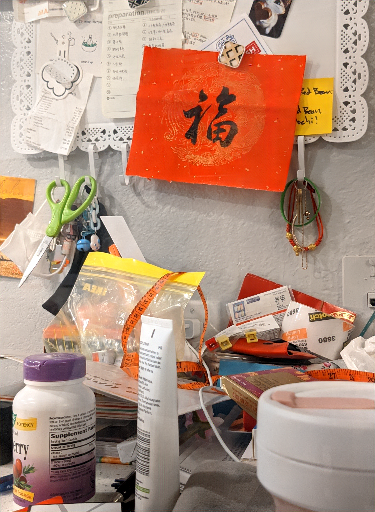}};
                \spy [every spy on node/.append style={line width=0.06cm}, spy connection path={\draw[line width=0.06cm] (tikzspyonnode) -- (tikzspyinnode);}] on (0.5,0.6) in node at (-0.6,-1.25);
                \node [align=center, fill=white, anchor=north west] at (-1.95,2.53) {\footnotesize PSNR: $22.74$};
            \end{tikzpicture}
        \\
            \footnotesize (a) Ours - Mono
        &
            \footnotesize (b) Ours
        \\
    \end{tabular}\vspace{-0.2cm}
	\caption{A result from our mono baseline (a) and our approach (b). They have a comparable PSNR, yet 19 out of 20 participants in a user study preferred the result of (b).}\vspace{-0.4cm}
	\label{fig:metric}
\end{figure}

\begin{figure*}\centering
    \setlength{\tabcolsep}{0.05cm}
    \setlength{\itemwidth}{2.35cm}
    \hspace*{-\tabcolsep}\begin{tabular}{ccccccc}
            \begin{tikzpicture}[spy using outlines={3787CF, magnification=5, size=\itemwidth, connect spies,
    every spy in node/.append style={line width=0.06cm}}]
                \node [inner sep=0.0cm] {\includegraphics[width=\itemwidth, trim={0.0cm 0.0cm 0.0cm 2.0cm}, clip]{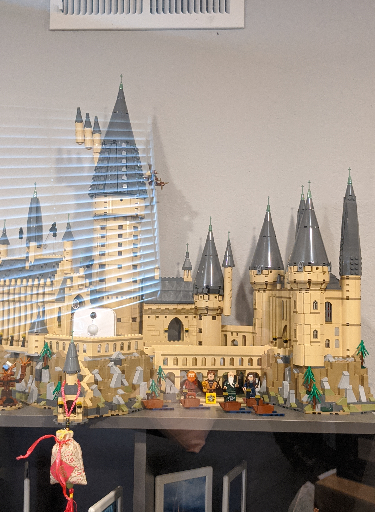}};
                \spy [every spy on node/.append style={line width=0.06cm}, spy connection path={\draw[line width=0.06cm] (tikzspyonnode) -- (tikzspyinnode);}] on (-0.9,0.75) in node at (0.0,2.7);
                \spy [every spy on node/.append style={line width=0.06cm}, spy connection path={\draw[line width=0.06cm] (tikzspyonnode) -- (tikzspyinnode);}] on (0.0,-0.87) in node at (0.0,-2.7);
            \end{tikzpicture}
        &
            \begin{tikzpicture}[spy using outlines={3787CF, magnification=5, size=\itemwidth, connect spies,
    every spy in node/.append style={line width=0.06cm}}]
                \node [inner sep=0.0cm] {\includegraphics[width=\itemwidth, trim={0.0cm 0.0cm 0.0cm 2.0cm}, clip]{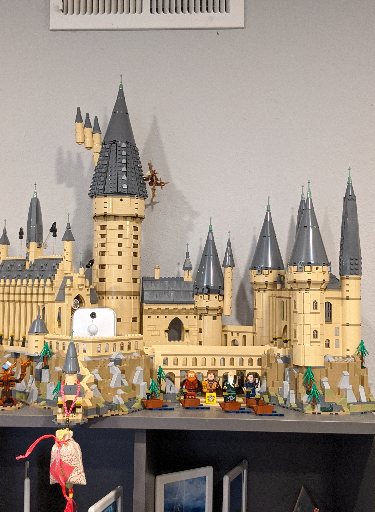}};
                \spy [every spy on node/.append style={line width=0.06cm}, spy connection path={\draw[line width=0.06cm] (tikzspyonnode) -- (tikzspyinnode);}] on (-0.9,0.75) in node at (0.0,2.7);
                \spy [every spy on node/.append style={line width=0.06cm}, spy connection path={\draw[line width=0.06cm] (tikzspyonnode) -- (tikzspyinnode);}] on (0.0,-0.87) in node at (0.0,-2.7);
            \end{tikzpicture}
        &
            \begin{tikzpicture}[spy using outlines={3787CF, magnification=5, size=\itemwidth, connect spies,
    every spy in node/.append style={line width=0.06cm}}]
                \node [inner sep=0.0cm] {\includegraphics[width=\itemwidth, trim={0.0cm 0.0cm 0.0cm 2.0cm}, clip]{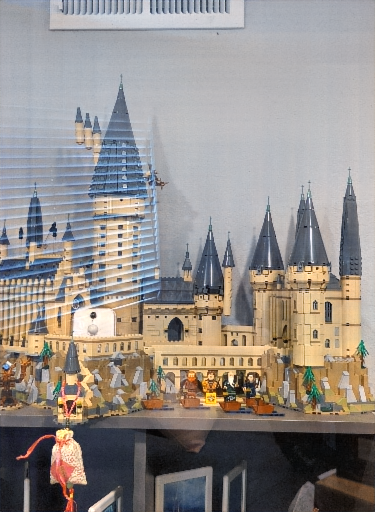}};
                \spy [every spy on node/.append style={line width=0.06cm}, spy connection path={\draw[line width=0.06cm] (tikzspyonnode) -- (tikzspyinnode);}] on (-0.9,0.75) in node at (0.0,2.7);
                \spy [every spy on node/.append style={line width=0.06cm}, spy connection path={\draw[line width=0.06cm] (tikzspyonnode) -- (tikzspyinnode);}] on (0.0,-0.87) in node at (0.0,-2.7);
            \end{tikzpicture}
        &
            \begin{tikzpicture}[spy using outlines={3787CF, magnification=5, size=\itemwidth, connect spies,
    every spy in node/.append style={line width=0.06cm}}]
                \node [inner sep=0.0cm] {\includegraphics[width=\itemwidth, trim={0.0cm 0.0cm 0.0cm 2.0cm}, clip]{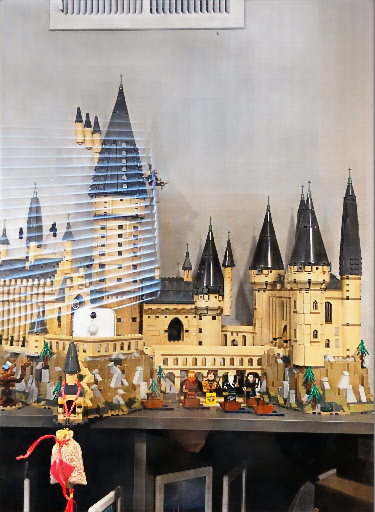}};
                \spy [every spy on node/.append style={line width=0.06cm}, spy connection path={\draw[line width=0.06cm] (tikzspyonnode) -- (tikzspyinnode);}] on (-0.9,0.75) in node at (0.0,2.7);
                \spy [every spy on node/.append style={line width=0.06cm}, spy connection path={\draw[line width=0.06cm] (tikzspyonnode) -- (tikzspyinnode);}] on (0.0,-0.87) in node at (0.0,-2.7);
            \end{tikzpicture}
        &
            \begin{tikzpicture}[spy using outlines={3787CF, magnification=5, size=\itemwidth, connect spies,
    every spy in node/.append style={line width=0.06cm}}]
                \node [inner sep=0.0cm] {\includegraphics[width=\itemwidth, trim={0.0cm 0.0cm 0.0cm 2.0cm}, clip]{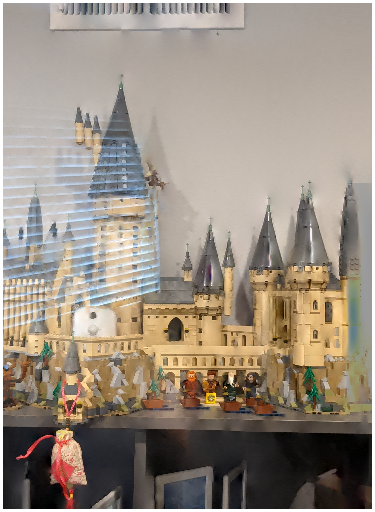}};
                \spy [every spy on node/.append style={line width=0.06cm}, spy connection path={\draw[line width=0.06cm] (tikzspyonnode) -- (tikzspyinnode);}] on (-0.9,0.75) in node at (0.0,2.7);
                \spy [every spy on node/.append style={line width=0.06cm}, spy connection path={\draw[line width=0.06cm] (tikzspyonnode) -- (tikzspyinnode);}] on (0.0,-0.87) in node at (0.0,-2.7);
            \end{tikzpicture}
        &
            \begin{tikzpicture}[spy using outlines={3787CF, magnification=5, size=\itemwidth, connect spies,
    every spy in node/.append style={line width=0.06cm}}]
                \node [inner sep=0.0cm] {\includegraphics[width=\itemwidth, trim={0.0cm 0.0cm 0.0cm 2.0cm}, clip]{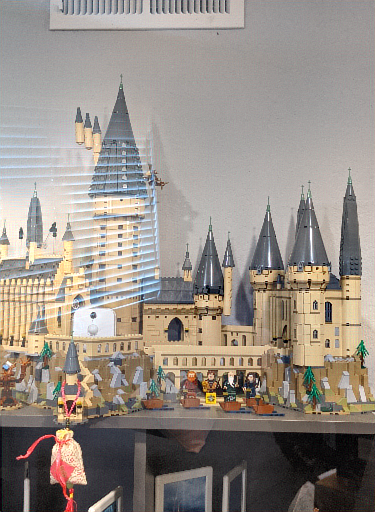}};
                \spy [every spy on node/.append style={line width=0.06cm}, spy connection path={\draw[line width=0.06cm] (tikzspyonnode) -- (tikzspyinnode);}] on (-0.9,0.75) in node at (0.0,2.7);
                \spy [every spy on node/.append style={line width=0.06cm}, spy connection path={\draw[line width=0.06cm] (tikzspyonnode) -- (tikzspyinnode);}] on (0.0,-0.87) in node at (0.0,-2.7);
            \end{tikzpicture}
        &
            \begin{tikzpicture}[spy using outlines={3787CF, magnification=5, size=\itemwidth, connect spies,
    every spy in node/.append style={line width=0.06cm}}]
                \node [inner sep=0.0cm] {\includegraphics[width=\itemwidth, trim={0.0cm 0.0cm 0.0cm 2.0cm}, clip]{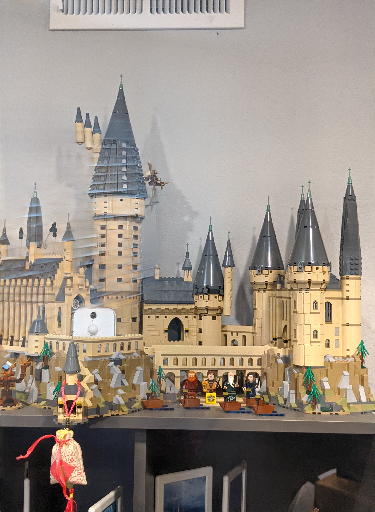}};
                \spy [every spy on node/.append style={line width=0.06cm}, spy connection path={\draw[line width=0.06cm] (tikzspyonnode) -- (tikzspyinnode);}] on (-0.9,0.75) in node at (0.0,2.7);
                \spy [every spy on node/.append style={line width=0.06cm}, spy connection path={\draw[line width=0.06cm] (tikzspyonnode) -- (tikzspyinnode);}] on (0.0,-0.87) in node at (0.0,-2.7);
            \end{tikzpicture}
        \\
            \footnotesize Input
        &
            \footnotesize Truth
        &
            \footnotesize Zhang~\etal~\cite{Xuzhang_CVPR_2018}
        &
            \footnotesize Wen~\etal~\cite{Wen_CVPR_2019}
        &
            \footnotesize Li \& Brown~\cite{Li_ICCV_2013}
        &
            \footnotesize Ours - Mono
        &
            \footnotesize Ours
        \\
    \end{tabular}\vspace{-0.2cm}
    \caption{Qualitative comparison. Please see the supplementary material for a tool-supported visual comparison.}\vspace{-0.4cm}
    \label{fig:qualitative}
\end{figure*}

\vspace{0.05in}
\noindent\textbf{Ablation:} We analyzed different components of our proposed network composition in an ablation study and tried four variations: 1) ``Zhang-like'', i.e., training the model from Zhang~\etal~\cite{Xuzhang_CVPR_2018} on our dataset, 2) ``Mono'', by only using a single input, 3) ``Concat'', by concatenating the input images without explicitly aligning them first, and 4) ``Regular Flow'', by replacing the flow network with the one trained on images without reflection. Table~\ref{tbl:ablation} shows the quantitative results. ``Mono'' outperforms ``Zhang-like'', which shows that the GridNet network architecture is well suited to this task. Also, our network with reflection invariant flow outperforms both ``Concat'' and ``Regular Flow''. This exemplifies the importance of reflection-invariant alignment. 

\vspace{0.05in}
\noindent\textbf{Quantitative:} The quantitative comparison of the recovered transmission image is shown in Table~\ref{tbl:quantstudy}, it includes comparisons to four baseline algorithms: two single-frame reflection removal algorithms by Zhang~\etal~\cite{Xuzhang_CVPR_2018} and Wen~\etal~\cite{Wen_CVPR_2019}, one multi-frame algorithm by Li and Brown~\cite{Li_ICCV_2013}, and a single-image ablation of our approach (``Ours - Mono''). Our proposed dual-view approach outperforms all baselines on all metrics, demonstrating the effectiveness of our method. However, using the input image itself as a baseline already shows surprisingly good results, especially on the real-world test dataset. This raises the question of whether or not traditional quality metrics are suitable for evaluating reflection removal. This is exemplified by Figure~\ref{fig:metric}, which shows example results with similar PSNR but a strong preference by human examiners for one over the other. We thus subsequently further compare the results though a user study.

\noindent\textbf{User study:} We conducted an A/B user study with $20$ participants that were not related to this project, including $2$ professional photographers, to further evaluate our results. We chose subsets for each test set to keep the number of comparisons for each participant below $200$. For our rendered test set, we chose $3$ challenging samples from each virtual test world resulting in $9$ images. For our real-world test set, we chose the center and right cameras from the first capture in each set, resulting in $20$ images. We asked each participant to select ``the best looking images''. The results of this are included in Table~\ref{tbl:quantstudy}. Overall, our approach is preferred over the baselines in the vast majority of cases.

\vspace{0.05in}
\noindent\textbf{Qualitative:} We show a representative example result in Figure~\ref{fig:qualitative}, which shows that our proposed dual-view approach can better remove challenging reflections in our test data. Please also consider the supplementary material for a comparison tool which includes many more examples.

\begin{figure}\centering
    \setlength{\tabcolsep}{0.05cm}
    \setlength{\itemwidth}{4.11cm}
    \hspace*{-\tabcolsep}\begin{tabular}{cc}
            \begin{tikzpicture}
                \definecolor{arrowcolor}{RGB}{238,127,14}
                \node [anchor=south west, inner sep=0.0cm] (image) at (0,0) {
                    \includegraphics[width=\itemwidth, trim={0.0cm 1.0cm 0.0cm 0.0cm}, clip]{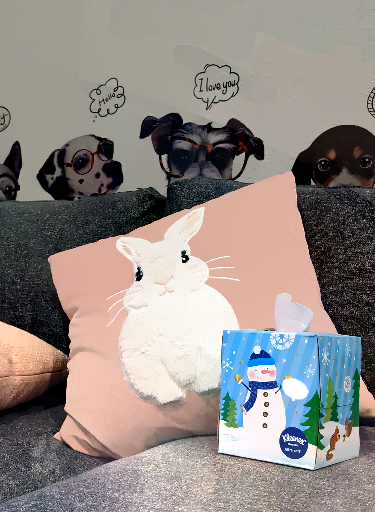}
                };
                \begin{scope}[x={(image.south east)},y={(image.north west)}]
                    \draw [double arrow=0.2cm with white and arrowcolor] (0.13,0.9) -- (0.41,0.87);
                    \draw [double arrow=0.2cm with white and arrowcolor] (0.1,0.7) -- (0.2,0.5);
                    \draw [double arrow=0.2cm with white and arrowcolor] (0.44,0.07) -- (0.7,0.17);
                \end{scope}
            \end{tikzpicture}
            &
            \begin{tikzpicture}
                \definecolor{arrowcolor}{RGB}{97,157,71}
                \node [anchor=south west, inner sep=0.0cm] (image) at (0,0) {
                    \includegraphics[width=\itemwidth, trim={0.0cm 1.0cm 0.0cm 0.0cm}, clip]{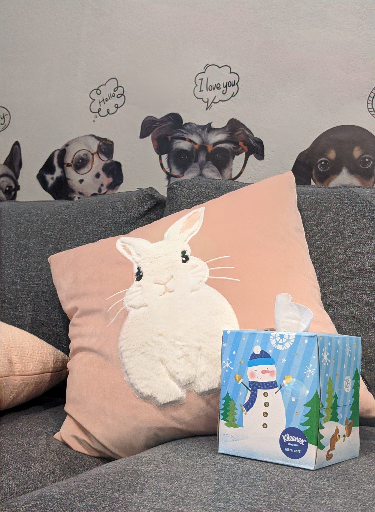}
                };
                \begin{scope}[x={(image.south east)},y={(image.north west)}]
                    \draw [double arrow=0.2cm with white and arrowcolor] (0.13,0.9) -- (0.41,0.87);
                    \draw [double arrow=0.2cm with white and arrowcolor] (0.1,0.7) -- (0.2,0.5);
                    \draw [double arrow=0.2cm with white and arrowcolor] (0.44,0.07) -- (0.7,0.17);
                \end{scope}
            \end{tikzpicture}
        \\
            \footnotesize (a) Dual Pixels~\cite{Punnappurath_CVPR_2019}
        &
            \footnotesize (b) Ours
        \\
    \end{tabular}\vspace{-0.2cm}
	\caption{On our stereo data, the recent dual-pixel technique~\cite{Punnappurath_CVPR_2019} flattens textures and does not catch all reflections.}\vspace{-0.4cm}
	\label{fig:dualpixel}
\end{figure}

\subsection{Dual-Pixel Reflection Removal} 

Recently, Punnappurath~\etal~\cite{Punnappurath_CVPR_2019} proposed a dual-pixel reflection removal technique. Dual-pixel images superficially resemble stereo pairs in that they both capture two perspectives of a scene. However, this dual-pixel technique performs poorly when applied to our stereo data: it achieved a PSNR/SSIM/LPIPS score of $17.82$/$0.774$/$0.230$ on our rendered test set and $14.52$/$0.567$/$0.350$ on our real-world test set (examples shown in Figure~\ref{fig:dualpixel}). This is consistent with recent work on dual-pixel imagery for depth estimation~\cite{Garg_ICCV_2019}, which has shown that dual-pixel footage is sufficiently different from stereo in terms of photometric properties that it benefits from being treated as a distinct problem domain.